\documentclass[
twocolumn,
hf
]{ceurart}

\usepackage{enumitem}
\usepackage{array}
\usepackage{stfloats}

\begin{document}

\copyrightyear{2022}
\copyrightclause{Copyright for this paper by its authors.
  Use permitted under Creative Commons License Attribution 4.0
  International (CC BY 4.0).}

\conference{Swisstext'22: Swiss Text Analytics Conference,
  June 08--10, 2022, Lugano, Switzerland}

\title{Keyword Extraction in Scientific Documents}


\author[1,2,8]{Susie Xi Rao}[
email=srao@ethz.ch,
url=susierao.github.io,
type=author
]

\author[1,2,8]{Piriyakorn Piriyatamwong}[
email=ppiriyata@student.ethz.ch,
url=linkedin.com/in/ppiriya,
type=author
]

\author[3,8]{Parijat Ghoshal}[
email=parijat.ghoshal@nzz.ch,
url=linkedin.com/in/parijat-ghoshal,
type=author
]

\author[4]{Sara Nasirian}[
email=sara.nasirian@supsi.ch,
url=linkedin.com/in/sara-nasirian-phd,
type=author
]

\author[5]{Emmanuel de Salis}[
email=emmanuel.desalis@he-arc.ch,
url=he-arc.ch/ingenierie/groupes-de-competences/analyse-de-donnees,
type=author
]

\author[6]{Sandra Mitrovi\'{c}}[
email=sandra.mitrovic@idsia.ch,
url=nlp.idsia.ch/sandramitrovic.html,
type=author
]

\author[7]{Michael Wechner}[
email=michael.wechner@wyona.com,
url=wyona.com,
type=author
]

\author[7]{Vanya Brucker}[
email=vanya.brucker@wyona.com,
url=wyona.com,
type=author
]

\author[2]{Peter Egger}[
email=pegger@ethz.ch, 
url=cae.ethz.ch]

\author[1]{Ce Zhang}[%
email=ce.zhang@inf.ethz.ch,
url=ds3lab.inf.ethz.ch]

\address[1]{Systems Group, Department of Computer Science (ETH Zurich), Stampfenbachstrasse 114, 8092 Zurich, Switzerland}
\address[2]{Chair of Applied Economics, Department of Management, Technology, and Economics (ETH Zurich), Leonhardstrasse 21, 8092 Zurich, Switzerland}
\address[3]{Neue Z{\"u}rcher Zeitung AG (NZZ AG), Falkenstrasse 12, 8008 Zurich, Switzerland}
\address[4]{Information Systems and Networking Institute, Department of Innovative Technologies (SUPSI), Via la Santa 1, 6962 Lugano-Viganello, Switzerland}
\address[5]{Data Analytics Group, Department of Engineering (HE-Arc), Rue de la Serre 7, 2610 St-Imier, Switzerland}
\address[6]{Dalle Molle Institute for Artificial Intelligence USI-SUPSI, Department of Innovative Technologies (SUPSI), Via la Santa 1, 6962 Lugano-Viganello, Switzerland}
\address[7]{Wyona AG, Fritz-Fleiner Weg, 8044 Zurich, Switzerland}
\address[8]{These authors contributed equally to this work. }

\begin{abstract}
    The scientific publication output grows exponentially. Therefore, it is increasingly challenging to keep track of trends and changes.  Understanding scientific documents is an important step in downstream tasks such as knowledge graph building, text mining, and discipline classification. 
    In this workshop, we provide a better understanding of keyword and keyphrase extraction from the abstract of scientific publications. 
\end{abstract}


\maketitle

\section{Introduction}

Keyphrases are single- or multi-word expressions (often nouns) that capture the main ideas of a given text, but do not necessarily appear in the text itself \cite{10.1007/978-3-540-77094-7_41, kim-kan-2009-examining, DBLP:conf/ijcai/FrankPWGN99}. Keyphrases have been shown to be useful for many tasks in the Natural Language Processing (NLP) domain, such as (1.) indexing, archiving and pinpointing information in the Information Retrieval (IR) domain \cite{DBLP:conf/ijcai/FrankPWGN99, GUTWIN199981, 10.5555/1364846.1364848, DBLP:journals/corr/abs-2109-05979}, (2.) document clustering \cite{DBLP:conf/ijcai/FrankPWGN99, 4427539, 10.1007/11510888_26}, and (3.) summarizing texts \cite{DBLP:conf/ijcai/FrankPWGN99, 10.5555/1613172.1613178, 10.4018/ijtd.2014040103, 10.1145/2980258.2980442}, just to name a few.




Keyphrase extraction has been at the forefront of various application domains, ranging from the scientific community \cite{10.1007/978-3-540-77094-7_41, kim-kan-2009-examining, kim-etal-2010-semeval}, finance \cite{10.1007/978-981-15-8458-9_12, su11051277}, law \cite{10.1007/978-3-319-10888-9_7}, news media \cite{10.1145/2980258.2980442, 2022arXiv220308118W, piskorski-etal-2021-exploring}, patenting \cite{suzuki-takatsuka-2016-extraction, e20020104}, and medicine \cite{10.1145/3473939, 7733254, LI2006668}. Despite being a seemingly straightforward task for human domain experts, performing automatic keyphrase extraction is a challenging task.

\paragraph{\textbf{Challenge 1: Benchmark Dataset and Keyword Reference List.}}

One main reason is the lack of benchmark datasets and keyword reference lists, as authors often do not provide their keyphrase list unless explicitly requested or required to do so \cite{ DBLP:conf/ijcai/FrankPWGN99}. In scientific publications, we see a large variation across domains (e.g., economics, computer science, mathematics, engineering fields, humanities). For instance, publications in some disciplines, such as economics, are required to have author-generated or journal-curated keywords, while in other domains, such as computer science and engineering fields, not all publication venues (e.g., journals, proceedings) require authors to input keywords. 

In less technical domains, such as news media, keyphrase lists may be more accessible in terms of the availability and the ease of manually curating the keyphrase list, even when reference lists are not readily available. This is because in the news domain, people have particular interests in Named Entities (labelled entities such as person, location, event, time), as we will discuss in Section~\ref{sec:sys3}. However, manually curating the keyphrase list in general is often practically infeasible--hiring domain experts is costly, while crowdsourcing the annotation is difficult to control the quality \cite{kim-kan-2009-examining, DBLP:conf/ijcai/FrankPWGN99, 10.1145/2980258.2980442}. 

With limited availability of benchmark datasets, large language models--which succeed in other NLP tasks--simply fail to optimize and generalize, as they generally require a large, well-annotated training dataset \cite{2022arXiv220308118W}. The lack of training datasets also poses challenges for the evaluation of keyword extraction systems. 
    
\begin{figure*}[!t] 
\begin{tabular}{cc}
    \includegraphics[width=0.45\linewidth]{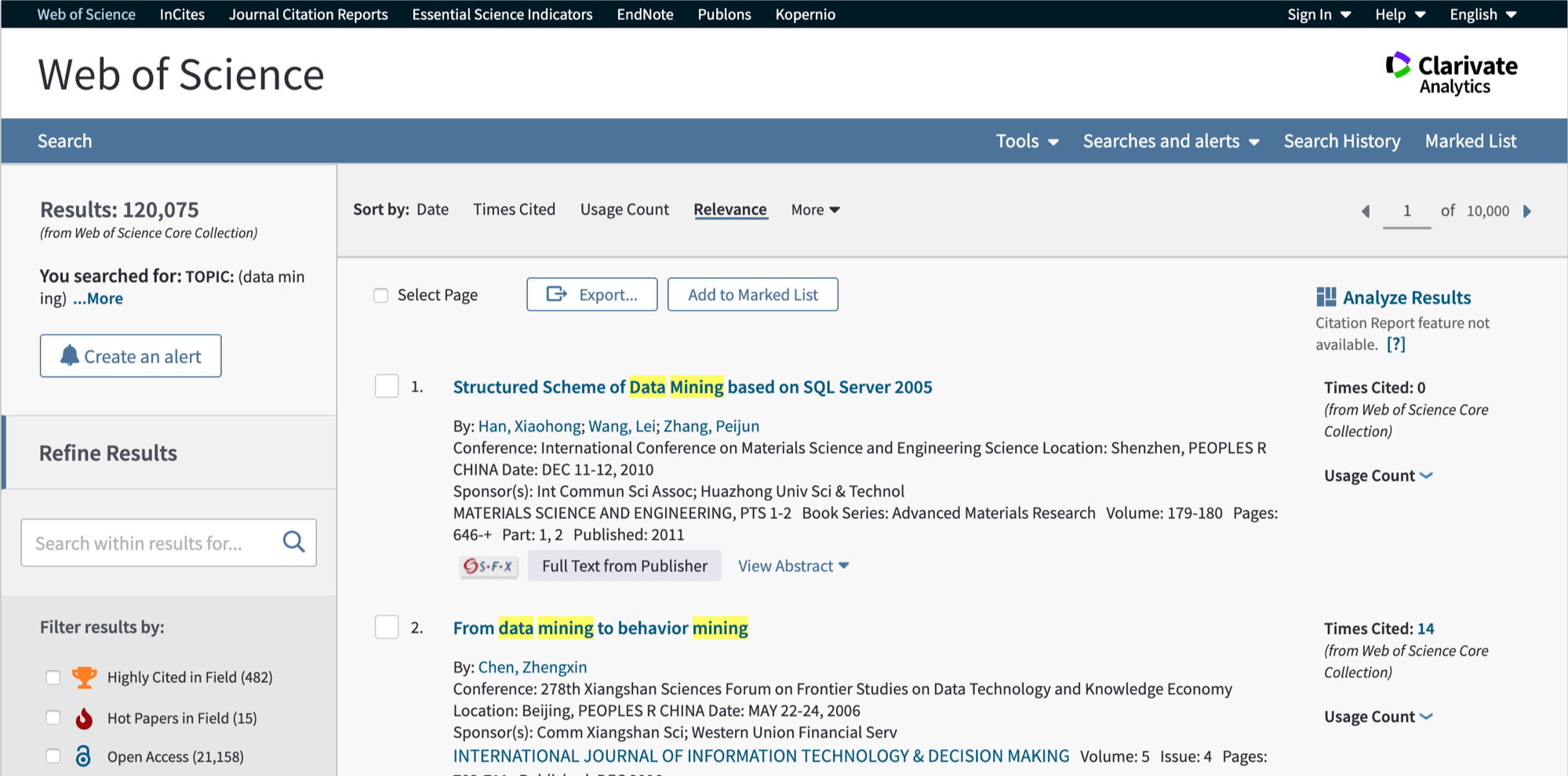}
   &  \includegraphics[width=0.45\linewidth]{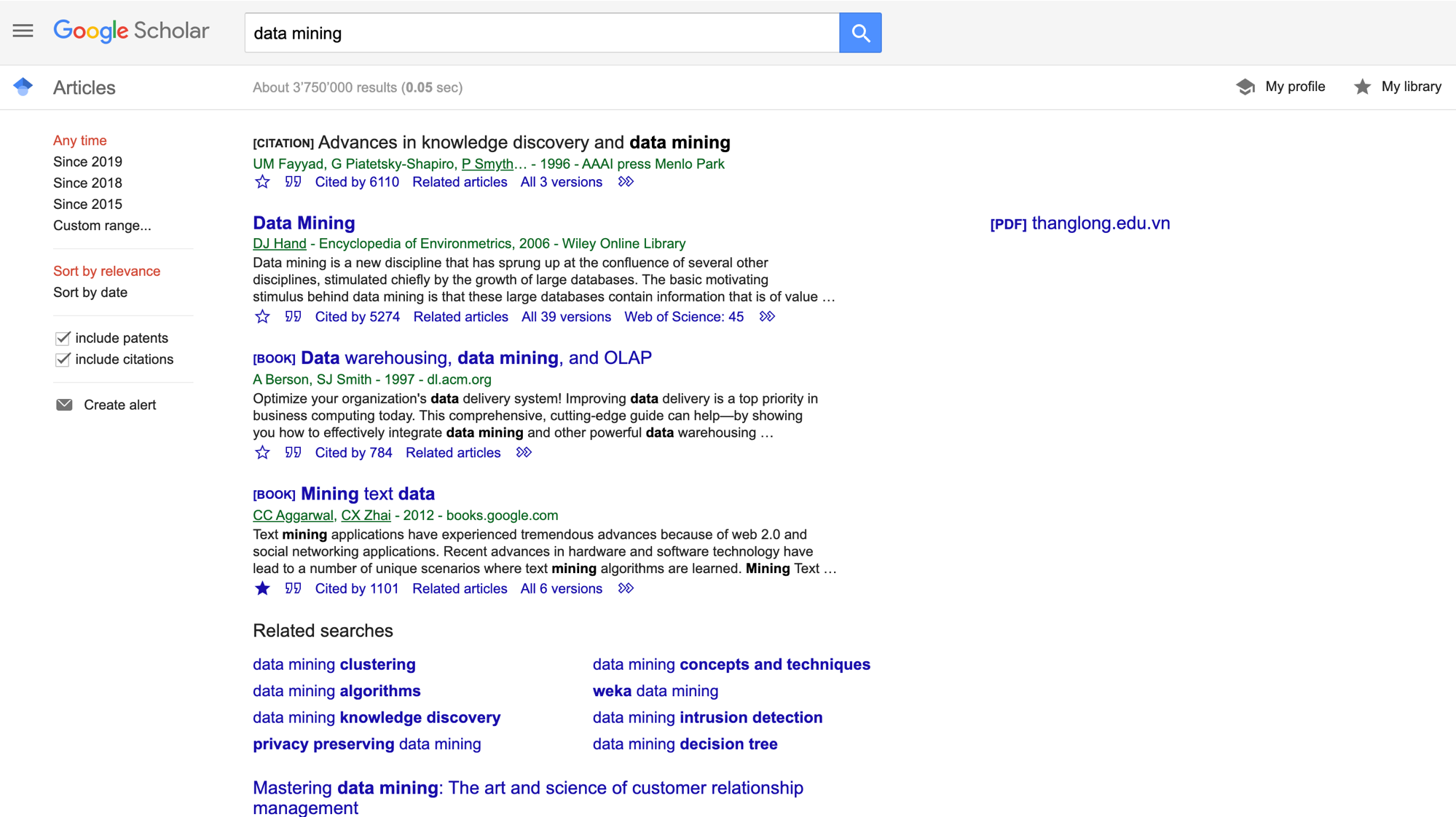} \\
   \textbf{(a) Web of Science.} & \textbf{(b) Google Scholar.} \\
   \includegraphics[width=0.45\linewidth]{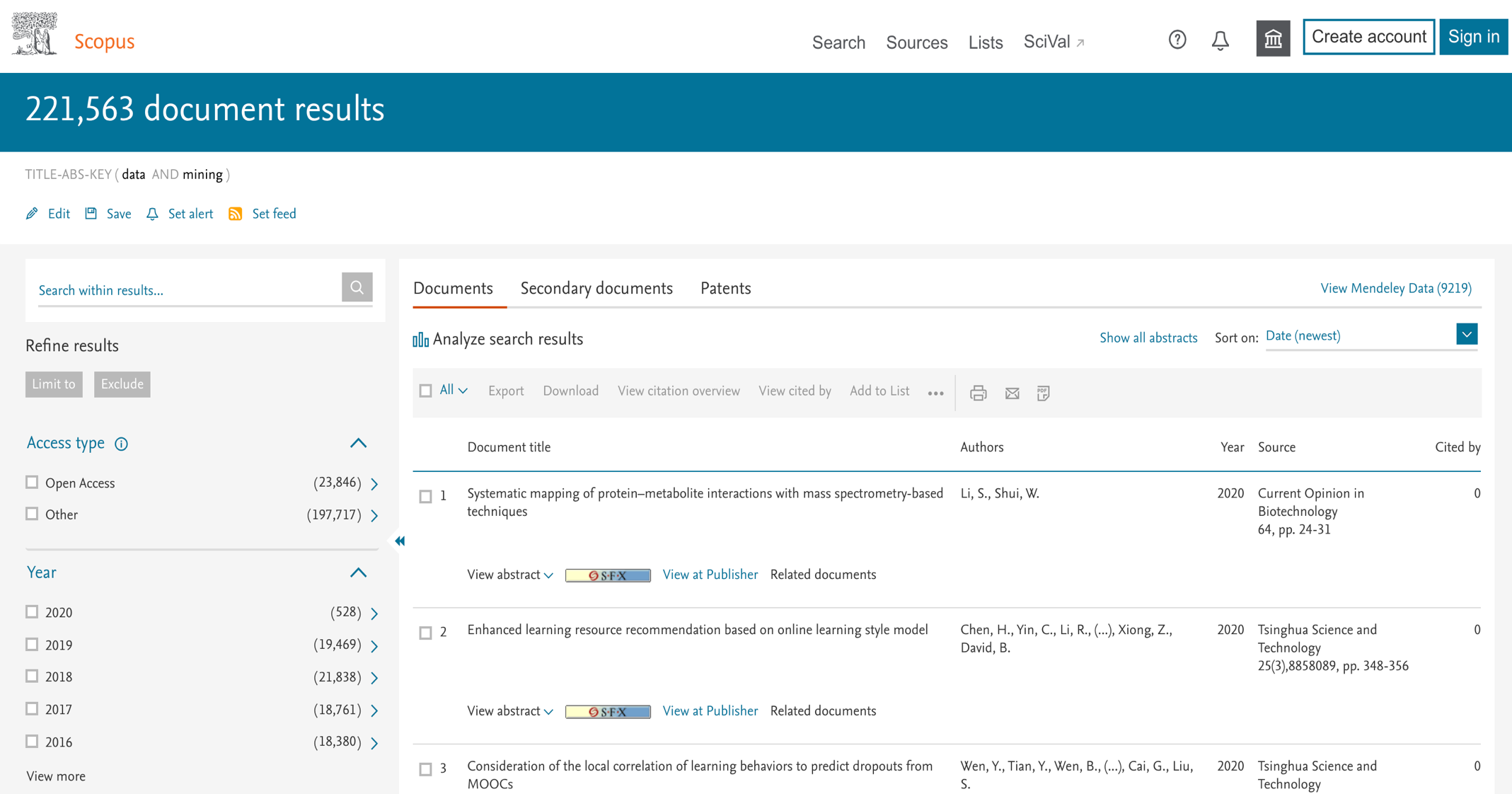}  & \includegraphics[width=0.45\linewidth]{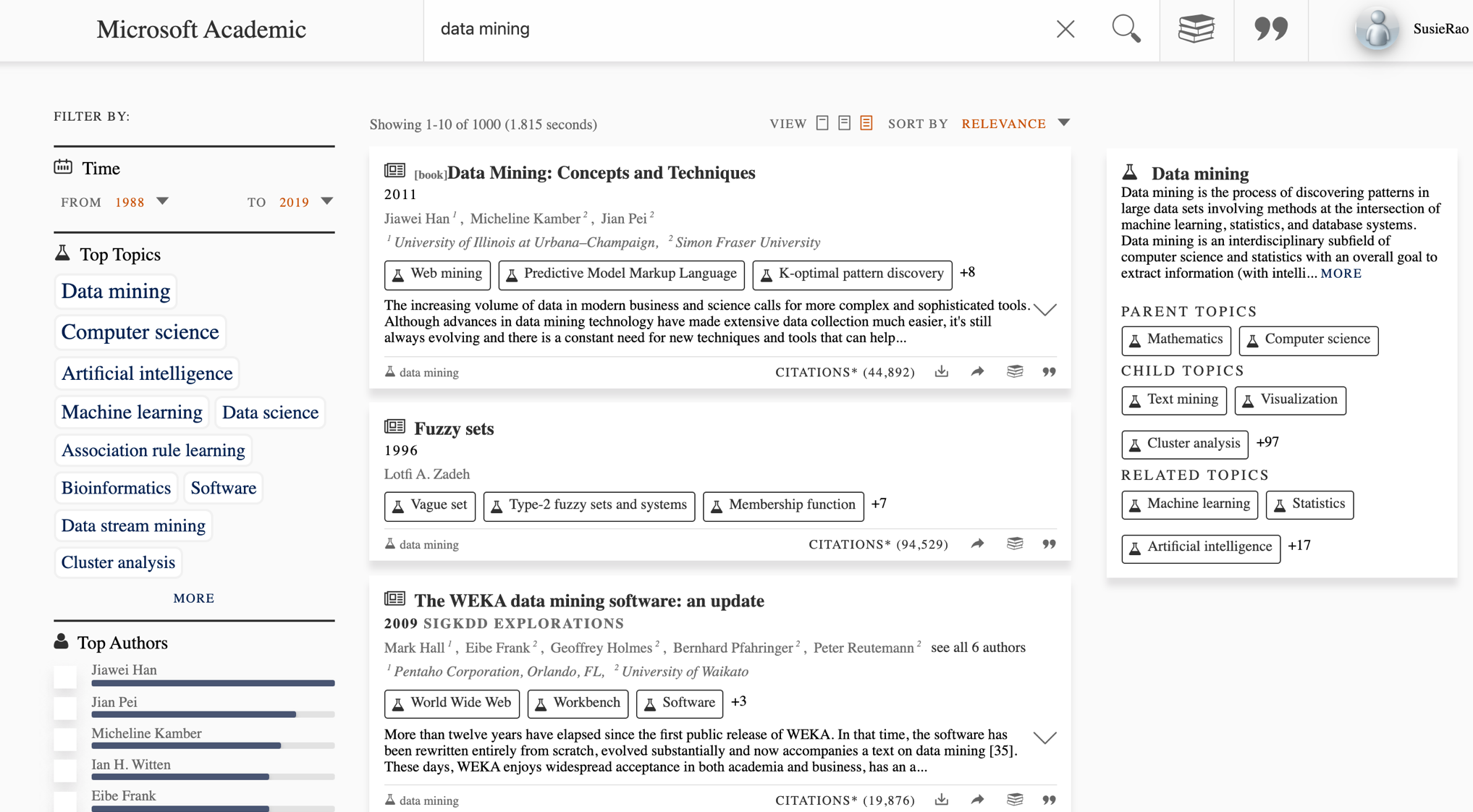} \\
   \textbf{(c) Scopus.} & \textbf{(d) Microsoft Academic.} \\
\end{tabular}
\caption{Comparison of various academic products with the query for ``data mining''.}
\label{fig:academic-products}
\end{figure*}

\paragraph{\textbf{Challenge 2: Evaluation of Keyword Extraction.}}

Defining an evaluation protocol and a corresponding metric is far from trivial for the following reasons. 
\begin{enumerate}[label=(\arabic*.), leftmargin=*]
    \item We should look at the ground truth list of keywords in a critical way. 
    As we mentioned above, there can exist more than one ground truth list of keyphrases given an abstract. The keyword list provided in our dataset is a reference list of words provided by authors or by publishers. One should only treat this list as a reference list, but not the one and only correct list of keywords. 
    
    \item There are different aims in extracting keyphrases in system design. As we will introduce in the rationale of designing the three systems in Section~\ref{sec:systems}, the systems are designed to tackle various problems and, therefore, are optimized for different use cases. System 1 uses a simple TextRank algorithm (see Section~\ref{sec:sys1}), which outputs the most prominent set of keyphrases/keywords; System 2 uses TextRank on top of a clustering algorithm (see Section~\ref{sec:sys2}), which is targeted at grouping similar articles and then learns from the cluster of articles; and System 3 uses pre-trained models and tools on Named-Entity Recognition (NER) (see Section~\ref{sec:sys3}), with a goal to fully utilize existing models and tools by only pre-processing the input and/or post-processing the output.
    
    \item There are different objective functions that we want to optimize. Precision, recall, accuracy, false positive rate, and false negative rate are among the most common performance metrics for various application scenarios \cite{zehtabsalmasi2021frake}. We might also consider the order of keyphrases, for example, as sorted by criteria such as frequency, TextRank score  \cite{sym12111864, 8396997}. In search engines, the hit rate is also an important metric \cite{9602870}. Furthermore, one can evaluate exact matches and fuzzy matches. Fuzzy matches can also be broken down into two types: ``partial'' matches and semantically equivalent matches \cite{10.1007/978-3-319-07857-1_40, sym12111923, sym12111864}. There are other evaluation methods which account for the ranks and orders in the extracted keywords, see \href{https://medium.com/gumgum-tech/exploring-different-keyword-extractors-evaluation-metrics-and-strategies-ef874d336773}{this Medium article} for inspiration \cite{sym12111864}.
\end{enumerate}

\paragraph{\textbf{Challenge 3: Growing Number of Scientific Publications.}}

During the last decades, the number of scientific publications has increased exponentially each year \cite{20.500.11850/106263}, making it increasingly challenging for researchers to keep track of trends and changes, even strictly in their own field of interest \cite{DBLP:conf/ijcai/FrankPWGN99, DBLP:conf/iconference/HuaS21}. This bolsters the need for automatic keyword extraction for the use case as a text recommendation and summarization system. The effect of increasing publications is clearly visible in major academic search engines such as Google Scholar, Web of Science, Scopus, and Microsoft Academics. In a simple query (``data mining''), three out of four failed to bring up relevant scientific publications that are prominent in the field and anticipated by human domain experts. 

See the query results in Figure ~\ref{fig:academic-products} of a keyword search ``data mining'' in different academic products. We can see that the search results in different products vary largely, and it could be difficult for readers to choose between the different results without having prior knowledge of the field. So far, only Microsoft Academic Services (Figure ~\ref{fig:academic-products} (d)) has returned relevant research results that point to the most influential author and work in the field of data mining. This is because Microsoft Academic Service has enabled a hierarchical discipline classification (indexed by keyphrases) that supports its users when reviewing the search results. 
In summary, without relevant and correct keyphrases, effective indexing and thus querying is not feasible.

\paragraph{\textbf{Challenge 4: Domain-Specific Keyword Extraction.}}

Another challenge in keyphrase extraction is its domain-specific nature. One case is when a keyphrase extractor trained in generic texts may miss out technical terms that do not look like usual keyword noun chunks, such as the chemical name ``C4H*Cl'' \cite{Krallinger2015CHEMDNERTD}. The issue arises from the tokenization step: a non-alphabetic character such as ``4'' and ``*'' might be treated as a separator, and thus such a keyword gets split into ``C'', ``H'' and ``Cl'', losing its original notion. Even if the separator works perfectly, this type of chemical name would still confuse keyphrase extractors that filter candidate keyphrase based on Part-of-Speech (POS) tags. This is because for POS-based extractors, it is unclear whether ``C4H*Cl'' is an adjective, a noun or other POS tags.

Another case is when the keyphrase consists of a mix of generic and specific words, such as ``Milky Way''. ``Way'' is generally a stopword \cite{10.5555/275537.275705}, so the keyphrase extractor might only be able to detect ``Milky'' and throw away ``Way'' without realizing that the term ``Way'' is not a stopword in this specific context. 

Finally, we would like to mention KeyBERT, a state-of-the-art BERT-based keyword extractor \cite{grootendorst2020keybert}. KeyBERT works by extracting multi-word chunks whose vector embeddings are most similar to the original sentence. Without considering the syntactic structure of the text, KeyBERT sometimes outputs keyphrases that are incorrectly trimmed, such as ``algorithm analyzes'', ``learning machine learning''. This problem only worsens with the aforementioned examples from chemistry and astronomy, since it is not straightforward how to tokenize, i.e., ``split'', words and how to handle non-alphabetic characters.


\paragraph{\textbf{Our Goals and Contributions in this Workshop.}}
Despite the challenges, keyphrase extraction is an important step for many downstream tasks, as already described. In this workshop, we aim to cover the foundations of keyphrase extraction in scientific documents and provide a discussion venue for academia and industries on the topic of keyword extraction. 
Our contributions in the workshop are as follows.

\begin{enumerate}[label=(\arabic*.), leftmargin=*]
    \item We make a new use of the existing dataset from the Web of Science (WOS) \cite{kowsari2017HDLTex}. This dataset has been used as a benchmark dataset for hierarchical classification systems. Since it comes with reference lists of keywords, we utilize it as a benchmark dataset for keyword extraction. In this workshop, together with the participants, we study the feasibility of that dataset in three systems.
    
    \item We introduce three commonly used systems in academia and industry for keyword extraction. For the various use cases of keyword extraction, we also design baseline evaluation metrics for each system. 
    
    \item We encourage participants to discuss, extend, and evaluate the systems that we have introduced.
\end{enumerate}

\paragraph{\textbf{System Design of Keyword Extraction.}}

For the keyword extraction, we provide two systems based on the unsupervised, graph-based algorithm TextRank \cite{mihalcea-tarau-2004-textrank}. System 1 (see Section~\ref{sec:sys1}) is to develop the TextRank keyword extractor from scratch in order to understand the reasoning behind it. System 2 (see Section~\ref{sec:sys2}) combines the TextRank algorithm with the K-Means clustering algorithm \cite{1056489, macqueen1967classification} to provide keyphrases for each specific field (``cluster''). In System 3 (see Section~\ref{sec:sys3}), we cover the NER task, where an entity in the sentence is identified as person, organization, and others from predefined categories. We will focus primarily on the biomedical domain using the state-of-the-art biomedical NER tool called HunFlair \cite{10.1093/bioinformatics/btab042}. We also provide some baseline NERs for participants to evaluate.

Beyond this workshop, the keyphrase extraction and NER methods we present are applicable to other text corpora, including media texts and legal texts; one only has to aware the domain-specific nature and properly adjust the algorithm pipeline. As such, we have linked the \href{https://scikit-learn.org/stable/datasets/real_world.html#newsgroups-dataset}{20 newsgroup text dataset} for the participants to try their keyphrase extraction system on.

\section{Benchmark Dataset}

We take a subset of 46,985 records from the Web of Science dataset (WOS). The original WOS dataset is provided by Kamran Kowsari in the \textit{HDLTex: Hierarchical Deep Learning for Text Classification} paper \cite{kowsari2017HDLTex}. The original data was provided in .txt format. 

For the ease of work, we have pre-processed the original data and store it into .csv dataframe format, which would be most compatible with our Python working setup. The final dataframe is in the format as in Table \ref{table:df}, where (1) each record corresponds to a single scientific document, and (2) has the following columns:

\begin{itemize}[leftmargin=*]
    \item  \texttt{Domain}: the domain the document belongs to,
    \item \texttt{area}: the sub-domain the document belongs to,
    \item \texttt{keywords}: the list of keyphrases provided by the authors, stored as a single string with separator ``;'',
    \item \texttt{Abstract}: the abstract of the document.
\end{itemize}

Columns \texttt{Y1} and \texttt{Y2} which are simply the index of column \texttt{Domain} and \texttt{area}, respectively. Column \texttt{Y} are the sub-sub-domain, which we do not use here but includes for reference.

In the corpus, we are provided with scientific articles from seven domains: Medical, Computer Science (CS), Biochemistry, Psychology, Civil, Electronics and Communication Engineering (ECE), and Mechanical and Aerospace Engineering (MAE). Therefore, column \texttt{Y1} consists of unique values from 0 to 6.

\begin{table*}[h!]
\centering
\resizebox{1\linewidth}{!}{
\begin{tabular}{c c  c  c  c  c  c }
 \toprule
 \textbf{Y1} & \textbf{Y2} & \textbf{Y} & \textbf{Domain} & \textbf{area} & \textbf{keywords} & \textbf{Abstract} \\ 
 \midrule
 5 & 50 & 122 & Medical & Sports Injuries & Elastic therapeutic tape; Material properties; Tension test & The aim of this study was to analyze stabilometry in athletes... \\
 \midrule
 5 & 48 & 120 & Medical & Senior Health & Sports injury; Athletes; Postural stability & \shortstack{This study examined the influence of range of motion of \\ the ankle joints on elderly people’s balance ability...}  \\
 
 \bottomrule 
\end{tabular}}
\caption{A sample of the WOS benchmark dataset. }
\label{table:df}
\end{table*}

In Table \ref{table:df}, note that both records have the same domain \texttt{Y1} as ``5'' corresponding to \texttt{Domain} as ``Medical''. Their sub-domain \texttt{Y2} differs: the first record is about ``Sports Injuries'', while the second record is about ``Senior Health''. \texttt{keywords} and \texttt{Abstract} of each record match its sub-domain.

Finally, the records are splitted at the ratio 70:30 into the train/test sets with 32,899 and 14,096 abstracts, respectively. We provide the training set \textit{with} \texttt{keywords} column to the participants for the training of their keyword and/or NER extraction system, and the test set for the participants to evaluate the system. The reason for splitting the dataframe is so that the participants do not overfit their system towards the whole dataset. We encourage them to design their system based on the features learnt from the training set and apply the identical pipeline to the test set.

\section{Systems}
\label{sec:systems}
Now we discuss the three systems we provide to the participants as simple baselines for keyword extraction using the benchmark dataset. Certainly, there are various possible extensions to them. We list the participant contributions under Section~\ref{sec:contrib}.

\section{System 1: TextRank Algorithm}
\label{sec:sys1}

In System 1, we build the TextRank algorithm from scratch and add customizations to our needs, e.g., filtering by Part-of-Speech tags.

\subsection{TextRank}

The TextRank algorithm is a graph-based algorithm which, as the name suggests, is used to assign scores to texts, thereby giving a ranking \cite{mihalcea-tarau-2004-textrank}. It has numerous use cases in the NLP domain including webpage ranking (better known as PageRank), extractive text summarization, and keyword extraction \cite{mihalcea-tarau-2004-textrank, ART002312659, e20020104, piskorski-etal-2021-exploring, 10.1007/978-981-16-2377-6_46, 10.1145/3321408.3326659}. Across different use cases, the base TextRank algorithm remains the same; one only needs to adjust what is designated as nodes, edges, and edge weights when constructing the graph from the text corpus. The higher edge weight means the higher chance of choosing this particular edge to proceed to the next node. For example, in the web context, the PageRank Algorithm considers different webpages as nodes and the hyperlinks between webpage pairs as edges. Here, the edges are asymmetrically directed, since there could be a hyperlink from one page to another but not necessarily vice versa. The edges can then be weighted by the number of hyperlinks. 

In our keyword extraction, the TextRank algorithm  works by considering terms in text as graph nodes, term co-occurence as edges, and the number of co-occurence of two terms within a certain window as the edge weights. Note that the co-occurence window is a fixed pre-specified size (say, 5-gram within sentence boundary). Based on this notion, the graph is treated as weighted but undirected.

Subsequently, each term score is given by how ``likely'' an agent, starting at a random point in the graph and continuously jumping along the weighted edges, will end up at that term node after a long time horizon. The terms with higher scores are then considered more important, that is, the ``keywords'' extracted by the TextRank system.\footnote{In the web analogy, the webpage score would correspond to the chance that an Internet user would end up in that webpage after continuously browsing through the hyperlinks. In this sense, we retrieve the most popular webpages.} 

\subsection{Implementation}

We implement a very basic keyword extraction system based on the TextRank algorithm from scratch, in order for the participants to get hands-on experience on how the algorithm works. Subsequently, we propose additional improvement ideas so that participants have the opportunity to be creative and improve the basic system.

For implementation, we mainly use the Python package for natural language processing called \texttt{spaCy} \cite{ines_montani_2022_6621076}. spaCy utilizes pre-trained language models to perform many NLP tasks, among other things, Part-of-Speech tagging (PoS tagging), semantic dependency parsing, and Named-Entity Recognition. In our case, we use spaCy along with its small pre-trained model for English language (\texttt{en\_core\_web\_sm}) as a text pre-processor and tokenizer. The rest of tasks are handled by usual built-in Python libraries.

Our basic system consists of the following steps: 

\begin{enumerate}[label=(\arabic*.), leftmargin=*]
    \item Text pre-processing: stopword and punctuation removal.
    \item Text tokenization: tokenizing the text and build a vocabulary list.
    \item Build the adjacency matrix from the graph.
    \begin{itemize}[leftmargin=*]
        \item Matrix index in row and column: terms in the vocabulary list.
        \item Matrix entries: co-occurence of term pairs within the same window of pre-specified size.
    \end{itemize}
    \item Normalize the matrix and compute the stationary distribution of the matrix.
    \item Retrieve keyword(s) corresponding to terms with highest stationary probabilities.
\end{enumerate}

The implemented code is stored as a Jupyter notebook and hosted on \href{https://colab.research.google.com/drive/1nSqf\_UQaaJNm02PxOe4mCS6zij6HPPJb?usp=sharing}{Google Colaboratory} and allows the participant to test and work directly on the code online without local installation. There, the step-by-step description is provided and a code sanity check was performed. For example, our system extracts valid keywords ``cute'', ``dog'', ``cat'' (in descending order by term prominence) for a short text: ``This is a very cute dog. This is another cute cat. This dog and this cat are cute''.

\subsection{Further Ideas}

Inspired by existing keyword extraction systems in Python such as \texttt{summa} \cite{DBLP:journals/corr/BarriosLAW16} and \texttt{pke} \cite{boudin:2016:COLINGDEMO}, we have provided participants with a list of ideas to further improve the keyword extraction system along with hints for Python implementation using spaCy (see the \href{https://colab.research.google.com/drive/1nSqf\_UQaaJNm02PxOe4mCS6zij6HPPJb?usp=sharing}{Jupyter notebook}):

\begin{itemize}[leftmargin=*]
    \item Improve the pre-processing step:
    \begin{itemize}[leftmargin=*]
        \item Remove numbers.
        \item Standardize casings, such as lower-casing the entire text.
        \item Use a domain-specific or custom-made stopword list.
    \end{itemize}
    \item Improve the tokenization step: 
    \begin{itemize}[leftmargin=*]
        \item Filter by Part-of-Speech tags to only include nouns in the vocabulary list.
        \item Use a domain-specific tokenizer such as ScispaCy \cite{neumann-etal-2019-scispacy} for biomedical data. 
        \item Lemmatize or stem tokens before recording them in the vocabulary list and building the adjacency matrix, so that different versions of the same words (such as plural ``solitons'' and singular ``soliton'') are mapped to the same record.
    \end{itemize}
    \item Add the post-processing step: 
    \begin{itemize}[leftmargin=*]
        \item Exclude keywords that are too short.
    \end{itemize}
    \item Agglomerate keywords (and perhaps add back some stopwords) to form ``keyphrases'' (``the'' and ``of'' should not be removed within ``the Department of Health'').
\end{itemize}

Advanced participants are also directed to another Python package \texttt{NetworkX}, which has a built-in, computationally efficient implementation for the TextRank algorithm \cite{osti_960616}.

\subsection{Evaluation: Instance-Based Performance}
\label{sec:eval-sys1}
In System 1, the objective is \textit{instance-based}, that is, for each abstract, we need to evaluate how well the algorithm performs. The metric could be accuracy, that is, the ability to find as many keyphrases (compared to the reference list) as possible. We can also compute the precision and recall scores (micro or macro). We provide a simple baseline evaluation function in the \href{https://colab.research.google.com/drive/1HEm6gekQfh-EzNJQhx1wzrCab2dH--nr}{notebook}. Here, we allow fuzzy matching algorithms on the phrase level, where the cut-off ratio and the edit distance between the candidate term and the reference term can be adjusted. 


\section{System 2: TextRank with Clustering}
\label{sec:sys2}
In System 2, we extend the TextRank keyword extraction described in System 1 (see Section~\ref{sec:sys1}) and apply it to a group of texts clustered by the K-Means algorithm. In this way, we obtain a more focused keyword list specifically for each text group and learn about its characteristics.

\subsection{K-Means Algorithm}

The K-Means algorithm is a clustering algorithm which partitions points in a vector space into ``K'' clusters (``K'' being pre-specified), such that each point belongs to the cluster with the nearest cluster centroid (called ``Means'') \cite{1056489, macqueen1967classification}. It works in the following steps.

\begin{enumerate}[label=(\arabic*.), leftmargin=*]
    \item Assign k random points as the cluster ``means''.
    \item Doing the following until the convergence:
    \begin{enumerate}[leftmargin=*]
        \item Assignment step: Assign each point to the cluster with the least squared Euclidean distance to the cluster mean,
        \item Update step: Recalculate the ``mean'' as the average of all the points assigned to each cluster,
        \item Terminate when the cluster assignment stabilizes.
    \end{enumerate}
\end{enumerate}

We ultimately choose the K-Means algorithm for clustering because of its low complexity: it works very fast for large datasets like ours \cite{Xu2015, EZUGWU2022104743}. Often, one hidden caveat about the K-Means algorithm is the choice of the number of clusters ``K''. However, in our specific use case with the scientific publications, we usually have a good estimate based on the number of target disciplines. Therefore, K-Means serves our purpose well.

\subsection{Preprocessing: Sentence-BERT Embeddings}

As mentioned in the previous section, K-Means clusters points in a vector space. Therefore, we need to transform each text in our dataset into a vector representation. This is often done by averaging pre-trained word embeddings over all the words that appear in the document, regardless of whether they are context-free embeddings like GloVe \cite{pennington-etal-2014-glove} or contextualized embeddings like BERT \cite{devlin-etal-2019-bert}. However, this has been shown to perform worse than directly deriving contextualized sentence embeddings (Sentence-BERT \cite{reimers-gurevych-2019-sentence}). Therefore, we opt for contextualized sentence embeddings from Sentence-BERT, which is trained on the Siamese BERT networks \cite{reimers-gurevych-2019-sentence}. More technical details can be found in the original paper by N. Reimers and I. Gurevych \cite{reimers-gurevych-2019-sentence}.

The Sentence-BERT transforms each text into a 384-dimensional semantically meaningful vector, which is now ready to be an input to the K-Means algorithm for clustering. 

\subsection{Implementation}

We add the clustering step to our pipeline, which effectively results in the following procedure:

\begin{enumerate}[label=(\arabic*.)]
    \item For each document, extract its Sentence-BERT embedding,
    \item Cluster the documents into K groups based on their Sentence-BERT embeddings, i.e., by the sentence contents,
    \item For each document cluster, extract its keyphrases.
\end{enumerate}

First, we generate embedding representations for each text, which is very easy by the Python package \texttt{sentence-transformers}. The package \texttt{sentence-transformers} offers several pre-trained models for different purposes, from which we choose the small model (\texttt{all-MiniLM-L6-v2}).

Second, to group the documents, we use the implementation in the package \texttt{sklearn} \cite{scikit-learn}. Furthermore, we provide a cluster visualization using the package \texttt{matplotlib} \cite{Hunter:2007}. We set the parameter $K=7$ for the K-Means algorithm, which is the number of disciplines in the WOS dataset.

Finally, we extract the keyphrases from each cluster. Unlike in System 1, we do not implement the TextRank algorithm from scratch, but instead use the existing Python package \texttt{pke} \cite{boudin:2016:COLINGDEMO}. \texttt{pke} provides implementations of numerous keyword extraction algorithms from publications, as well as allowing customization such as Part-of-Speech tag filters and the limit on the maximum number of words in a single keyphrase. In our case, we simply use the basic TextRank algorithm, also to demonstrate that even the very basic algorithm can already yield satisfying outputs.

Like in System 1, the code implemented for System 2 is stored as a Jupyter notebook and hosted on \href{https://colab.research.google.com/drive/1tOVX6HJM51-nNO9sVSY73kRzn1a7oDPp?usp=sharing}{Google Colaboratory}. The step-by-step description is provided, and a code sanity check succeeds at characterizing a cluster: the cluster mostly consisting of medical articles has relevant keyphrases such as ``patient group'', ``treatment effects'', ``autism patient'' among the top-10 extracted keyphrase list.


\subsection{Further Ideas}

We invite participants to explore improvement ideas and provide coding hints on how to implement them on \texttt{pke}:

\begin{itemize}[leftmargin=*]
    \item Customize the TextRank algorithm:
    \begin{itemize}[leftmargin=*]
        \item Change the window size.
    \end{itemize}
    \item Use alternative keyword extraction algorithms to the TextRank algorithm, such as:
    \begin{itemize}[leftmargin=*]
        \item The TopicRank algorithm \cite{bougouin:hal-00917969},
        \item The Multipartite algorithm \cite{boudin-2018-unsupervised},
        \item The BERTopic algorithm
        \cite{grootendorst2022bertopic}.
    \end{itemize}
    \item Impose extra criteria on valid keyphrases, such as:
    \begin{itemize}[leftmargin=*]
        \item Change the maximum number of words allowed in a single keyphrase,
        \item Restrict the keyphrase to only contain the top certain percentage of all keywords.
    \end{itemize}
\end{itemize}

\subsection{Evaluation: Cluster-Based Performance}
\label{eval:sys2}
Using a similar evaluation function as in System 1 (See Section~\ref{sec:eval-sys1}), we now look at a \textit{cluster-based} objective. This means that we take all the keywords from the articles clustered in the same group and build a new reference list of keywords. Subsequently, the evaluation of the user-generated list will be compared with this expanded list. Notably, this approach increases the coverage of keywords in the reference, in the hope of covering more out-of-abstract keywords in this expanded list. However, it comes at the cost of increasing the denominator when we compare the user-generated list to the reference list. One way to better present the reference list of one cluster is to process the list by criteria such as frequency. Another way to evaluate is using word embedding similarities (c.f. KeyBert \cite{grootendorst2020keybert} as an example of leveraging embeddings). In this way, we have a better view of the extracted keywords and the degree to which the user-generated list is close to the reference list. In particular, this technique is useful for assessing the difference set between the user-generated list and the reference one. 

\section{System 3: Named-Entity Recognition as Keyword Extraction}
\label{sec:sys3}
The goal of system 3 is to emulate some of the constraints that may exist in a practical setting. These could be situations where a keyword extractor system cannot be implemented as the output of these systems may be incorrect or non-sensical. Another situation could be that one is required to use existing tools such as a Named-Entity Recognition system and must enact measures to improve the output of the model.

\subsection{Named Entities,  Named-Entity Recognition and Keyword Extraction}
A named entity (NE) in most cases is a proper noun, the most common categories being person, location and organization; however, other categories that are not proper nouns, such as temporal expressions, are also possible. Named-Entity Recognition consists of locating and classifying named entities mentioned in unstructured text into predefined categories~\cite[Chapter.~8.3]{jurafsky2018speech}. Keywords are single or multi-word expressions that under ideal circumstances should concisely represent the key content of a document~\cite[Page~3]{berry2010text}. As the goal of NER is to assign a label to spans of text~\cite[Chapter.~8.3]{jurafsky2018speech}, it is a classification task that can be solved by building a machine learning model \cite{mansouri2008named}. 

The difference between keyword extraction and NER is as follows. Named entities are words or phrases with a specific label determined by predefined classes of a given NER model. Therefore, these entities may not necessarily represent the essential content of a document. Keywords are not limited by the fixed categories of an NER model, and may contain named entities if those entities are representative of a given document. For example, a document about Heathrow Airport can contain keywords such as ``arrival'', ``customs'', ``departure'', ``duty free'', ``immigration'' and ``London''. Depending on the model classes, an NER model on the same text could extract entities such as ``British Airways'' (ORG), ``London'' (LOC), ``United Kingdom'' (LOC), etc. In this example, there is overlap between the keywords and named entities; however, due to the defining characteristics of both approaches, there is a significant difference between the lists.

\begin{figure}[!t] 
\includegraphics[width=\linewidth]{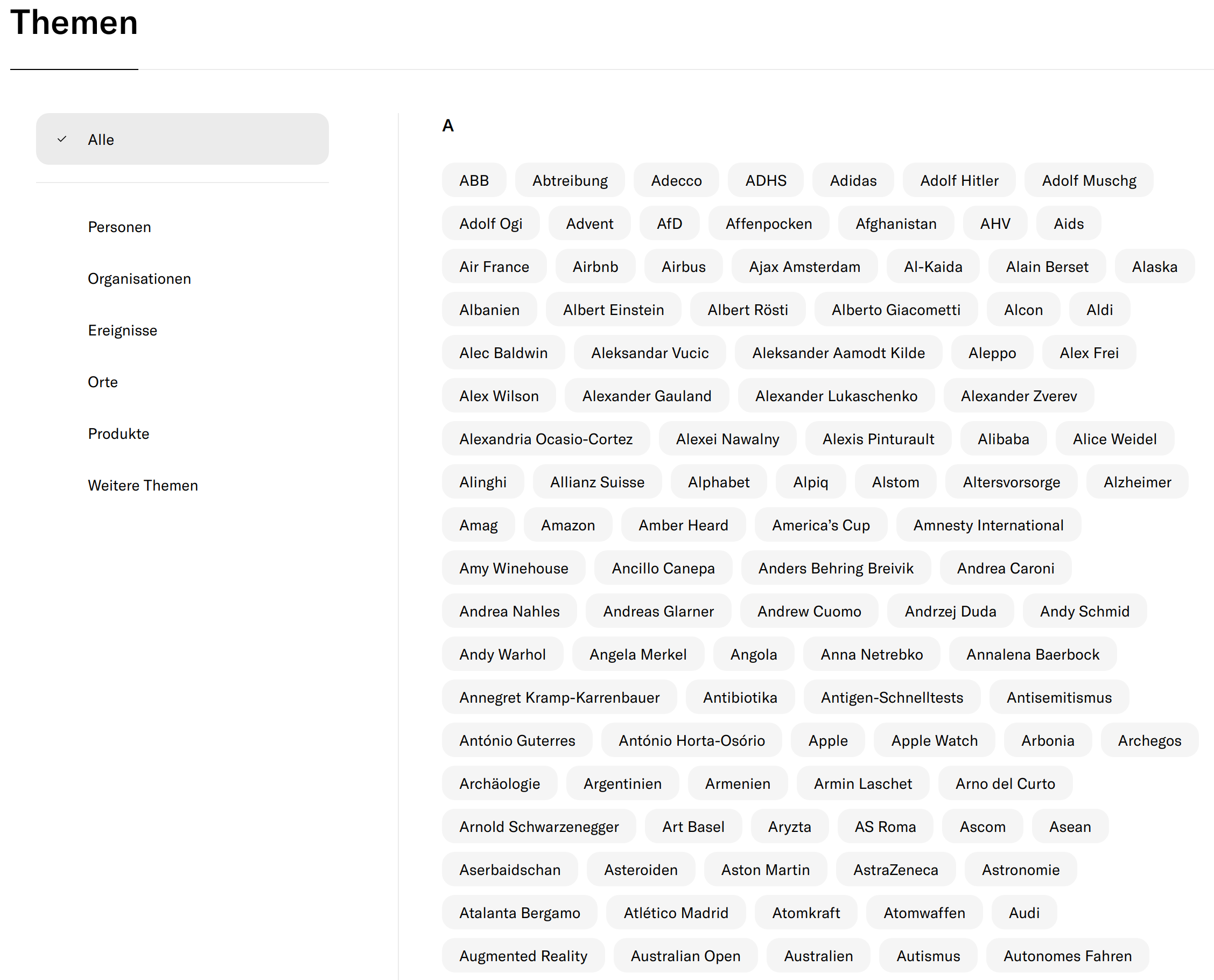}
\caption{NZZ Topic Page based on keywords and named entities from news articles. Accessible at \href{nzz.ch/themen}{nzz.ch/themen}.}
\label{fig:nzz-topic-page}
\end{figure}

Figure \ref{fig:nzz-topic-page} demonstrates the use of keyword extraction and named-entity recognition in the industry setting at Neue Zürcher Zeitung (NZZ), where key terms are extracted and relevant articles are assigned to the terms.

\subsection{Use of Keywords in the News Domain}
As mentioned above, for a given text, keywords and the output of a NER model may overlap. When it comes to analyzing news, a typical NER model (with common categories such as person, organization, and location) excels at finding named entities for the model-specific categories. However, only extracting the entities is inadequate for finding nuanced differences between multiple articles that contain identical named entities. In Table \ref{tab:nzz-article} we see the titles of 10 articles published in  Neue Zürcher Zeitung (NZZ) during March 2022. According to the NER model for German texts used internally by the NZZ, all articles have ``Ukraine'' (location) as a common named entity. Despite the similarities, there are thematic differences between these articles. After using a keyword extraction system that uses similar methodologies mentioned in Systems 1 and 2, keywords that are not named entities were found. These keywords demonstrate thematic groupings between the articles. The most common keyword for articles 1-4 is ``Flüchtlinge'' (``refugees''), and for articles 5-10 is ``Neutralität'' (``neutrality''). This difference can also be observed in the article titles, and upon closer inspection of the article content, it is evident that some of the articles (1-4) revolve around the topic of refugees from Ukraine, while other articles (5-10) discuss the notion of neutrality. Using named entities or, in some cases, a predefined list of keywords can be useful to define broad topic pages (see \href{https://www.nzz.ch/themen}{nzz.ch/themen}), but keywords offer concise yet semantically insights into the content of a document. Therefore, they can be potentially used to automatically identify possible subtopics with a news story or discover emerging topics from newly published articles.

\begin{table}[t]
\centering
\resizebox{1\linewidth}{!}{
\begin{tabular}{c | l }
 \toprule
 \textbf{Number} & \textbf{NZZ Article Title} \\ [0.5ex]
 \midrule
 1 & Eine Zürcherin nimmt ukrainische Flüchtlinge auf – und fühlt sich vom Staat alleingelassen \\
 2 & \shortstack{«Eine Solidaritätsbekundung auf Instagram zu posten, reicht nicht»: \\ Viele Zürcherinnen und Zürcher möchten Flüchtlinge aus der Ukraine bei sich zu Hause aufnehmen} \\
 3 & 150 Ukraine-Flüchtlinge sind im Kinderdorf – wie geht es weiter? \\
 4 & Krieg in der Ukraine: Wie ein SVP-Dorf Flüchtlinge aufnimmt \\
 5 & Neutralität im Ukraine-Krieg - wo genau  steht die Schweiz? \\
 6 & Neutralität: Fand in der Schweiz gerade eine Zeitenwende statt \\
 7 & Putin, die Schweiz und die zwei Seiten der Neutralität \\
 8 & Christoph Blocher: Neutralität ist nicht nur Selbstzweck \\
 9 & Sicherheitspolitik: Militärische Neutralität weiterdenken \\
 10 & Sicherheitspolitik: Solidarische Neutralität \\
 [1ex]
 \bottomrule
\end{tabular}
}
\caption{Titles of 10 articles published in  Neue Zürcher Zeitung (NZZ) during March 2022.}
\label{tab:nzz-article}
\end{table}

\subsection{Data Preparation}
\label{sec:data-sys3}
The FLAIR framework~\cite{akbik2019flair} was chosen as it contains many out-of-the-box NER models for generic and biomedical texts. Furthermore, the framework is also useful for integrating pre-trained embeddings and models. As many of the texts are from the biomedical domain, the ScispaCy library was used for word and sentence tokenization~\cite{DBLP:journals/corr/abs-1902-07669}. 
The results of the NER models were given to the participants. The \textit{ner-english} model is a 4-class NER model for English, which comes with FLAIR~\cite{akbik2018coling}. This model has the following categories: locations (LOC), persons (PER), organizations (ORG), and miscellaneous (MISC) \cite{tjong-kim-sang-de-meulder-2003-introduction}. 
We also provided participants with NER results from HunFlair \cite{10.1093/bioinformatics/btab042}, which is an NER tagger for biomedical texts. This biomedical NER tagger is based on the HUNER tagger, and has the follwing named-entity categories: Chemicals, Diseases, Species, Genes or Proteins, and Cell lines \cite{10.1093/bioinformatics/btz528}.  As an additional hint to participants, document embeddings for each item in the train and test sets, as well as word embeddings for the entire corpus, were generated from a fastText model\footnote{\url{https://fasttext.cc/} (last accessed: June 20, 2022).} trained on the English Common Crawl dataset (\texttt{cc.en.300.bin})\footnote{\url{https://fasttext.cc/docs/en/crawl-vectors.html} (last accessed: June 20, 2022).}. 

\subsection{Pre-Trained NER Models}
There are some disadvantages to using pre-trained NER models. One should take into consideration that using a pre-trained model to extract named entities out of documents from different domains can result in a fall in model performance \cite{MARRERO2013482}. The training data and categories of the model will influence the output. For example, the string ``ATP'' can be labeled as an organization (e.g. Association of Tennis Professionals) by one model and  as a chemical (e.g. adenosine triphosphate) by a biomedical-NER model.   Creating an NER model for a specific type of entity requires the annotation of a corpus, which can be a significant expense and effort for the user \cite{MARRERO2013482}. 

\subsection{Further Ideas}
The challenge of this system lies in working with pre-calculated data from systems that cannot be influenced. The participants are provided with multiple tables with the output of two different NER systems, fastText document, and word vectors (see Section~\ref{sec:data-sys3}). In addition, they also have a table at their disposal to verify whether a keyword for a given document is present in the abstract and whether it was discovered by any of the NER models (with 100\% string matches). The intuition of System 3 is that given the resources (cost, time, hardware), one needs to come up with the best possible strategies to detect meaningful keywords.

\subsection{Evaluation: Instance-based Performance}
In addition to the pre-calculated data, the participants were also given evaluation functions to compare differences between their system NER model output and the keyword list that came with the documents. There are cases where an item from the curated keyword list does not contain the keyword in the abstract, or contains a partial or inflected form of the keyword.  The evaluation function contains a partial string matching sequence, where one can choose the amount of character similarity between two strings. For example, a document has the label ``radio frequency”, but the string ``radio frequencies” is present in the abstract and the inflected form was also found by one of the NER models. For this case, participants can set a string similarity value (e.g., 80\% similarity) to circumvent the issues caused by inflected forms, or partially mentioned forms (``radio frequency” vs.~``radio frequency scanner”). Using the resources at their disposal, participants must develop the best possible strategies to build a system that can detect the maximum number of relevant keywords.

\section{Participant Contributions}
\label{sec:contrib}

Our participants have further investigated keyphrase extractions in System 1 and provided valuable contributions to our proceedings. Their original theses can be founded at the following \href{https://drive.google.com/drive/folders/1qlzeZxWK4cCLKlnLylu9OdhyUXNKAAkM?usp=sharing}{Google Drive folder}.

The basic TextRank keyword extractor in System 1 has been extended to account for the following data pre-processing steps: (1) remove numbers; (2) restrict valid keywords to only nouns; (3) restrict valid keywords by imposing the minimum string length. The contribution can be found on the \href{https://drive.google.com/drive/folders/1LByC0g44PsSUTUAwjaE2MJyFyzpooIvQ?usp=sharing}{Google Drive folder}. 

Additionally, the evaluation system has been generalized to output numerical performance scores, allowing simpler comparisons of different keyword extractors. 
The contribution can be found on the \href{https://drive.google.com/drive/folders/10HDCwzTRreovlr4EiWh18H915y7m5BPe?usp=sharing}{Google Drive folder}. 

Finally, a comparison between the TextRank algorithm and further unsupervised keyphrase extraction methods has been provided. The limitation of TextRank is that it only considers the co-occurences of the word pair and not the semantical meanings, which may cause certain extracted ``frequent'' word pairs to either be irrelevant or under-represented. Therefore, an experiment has been performed using the \texttt{pke} library to compare the performance of the TextRank algorithm and several other unsupervised keyphrase extraction algorithms on the benchmark test dataset.
The contribution can be found on the \href{https://drive.google.com/drive/folders/1v8PNTxjmE2B8GwlNHhKzc79Yar-cL2jb?usp=sharing}{Google Drive folder}. 

Beyond the academic setting, the use of keyword extractions is demonstrated in the industry setting, where \href{wyona.com}{Wyona AG} utilizes keyword extractors in the working pipeline of the Q\&A Chatbot ``Katie''. The contribution can be found on the \href{https://drive.google.com/drive/folders/1FBxD5pVwBgyPE8bVAczpabPUVBt8BPxz?usp=sharing}{Google Drive folder}. 

\section{Conclusion}
In this workshop, we provided the background and baseline systems for keyword extraction, shared a benchmark dataset on scientific keyword extraction, and invited contributions from participants from industry and academia. The methodologies discussed can be extended to keyword extraction in other domains (e.g., legal and news).

\begin{acknowledgments}
The authors would like to thank the organizers from \href{https://www.swisstext.org/}{SwissText2022} for hosting our workshop. 
Peter Egger and the Chair of Applied Economics acknowledge the support of the Department of Management, Technology, and Economics at ETH Zurich. 
Ce Zhang and the DS3Lab gratefully acknowledge the support from the Swiss State Secretariat for Education, Research and Innovation (SERI) under contract number MB22.00036 (for European Research Council (ERC) Starting Grant TRIDENT 101042665), the Swiss National Science Foundation (Project Number 200021\_184628, and 197485), Innosuisse/SNF BRIDGE Discovery (Project Number 40B2-0\_187132), European Union Horizon 2020 Research and Innovation Programme (DAPHNE, 957407), Botnar Research Centre for Child Health, Swiss Data Science Center, Alibaba, Cisco, eBay, Google Focused Research Awards, Kuaishou Inc., Oracle Labs, Zurich Insurance, and the Department of Computer Science at ETH Zurich. 
We would like to thank Neue Zürcher Zeitung for collaborating on this project. 
\end{acknowledgments}

\pagebreak

\bibliography{sample-ceur}

\begin{thebibliography}{65}
\expandafter\ifx\csname natexlab\endcsname\relax\def\natexlab#1{#1}\fi
\providecommand{\url}[1]{\texttt{#1}}
\providecommand{\href}[2]{#2}
\providecommand{\path}[1]{#1}
\providecommand{\DOIprefix}{doi:}
\providecommand{\ArXivprefix}{arXiv:}
\providecommand{\URLprefix}{URL: }
\providecommand{\Pubmedprefix}{pmid:}
\providecommand{\doi}[1]{\href{http://dx.doi.org/#1}{\path{#1}}}
\providecommand{\Pubmed}[1]{\href{pmid:#1}{\path{#1}}}
\providecommand{\bibinfo}[2]{#2}
\ifx\xfnm\relax \def\xfnm[#1]{\unskip,\space#1}\fi
\bibitem[{Nguyen and Kan(2007)}]{10.1007/978-3-540-77094-7_41}
\bibinfo{author}{T.~D. Nguyen}, \bibinfo{author}{M.-Y. Kan},
\newblock \bibinfo{title}{Keyphrase extraction in scientific publications},
\newblock in: \bibinfo{editor}{D.~H.-L. Goh}, \bibinfo{editor}{T.~H. Cao},
  \bibinfo{editor}{I.~T. S{\o}lvberg}, \bibinfo{editor}{E.~Rasmussen} (Eds.),
  \bibinfo{booktitle}{Asian Digital Libraries. Looking Back 10 Years and
  Forging New Frontiers}, \bibinfo{publisher}{Springer Berlin Heidelberg},
  \bibinfo{address}{Berlin, Heidelberg}, \bibinfo{year}{2007}, pp.
  \bibinfo{pages}{317--326}.
\bibitem[{Kim and Kan(2009)}]{kim-kan-2009-examining}
\bibinfo{author}{S.~N. Kim}, \bibinfo{author}{M.-Y. Kan},
\newblock \bibinfo{title}{Re-examining automatic keyphrase extraction
  approaches in scientific articles},
\newblock in: \bibinfo{booktitle}{Proceedings of the Workshop on Multiword
  Expressions: Identification, Interpretation, Disambiguation and Applications
  ({MWE} 2009)}, \bibinfo{publisher}{Association for Computational
  Linguistics}, \bibinfo{address}{Singapore}, \bibinfo{year}{2009}, pp.
  \bibinfo{pages}{9--16}.
\bibitem[{Frank et~al.(1999)Frank, Paynter, Witten, Gutwin, and
  Nevill{-}Manning}]{DBLP:conf/ijcai/FrankPWGN99}
\bibinfo{author}{E.~Frank}, \bibinfo{author}{G.~W. Paynter},
  \bibinfo{author}{I.~H. Witten}, \bibinfo{author}{C.~Gutwin},
  \bibinfo{author}{C.~G. Nevill{-}Manning},
\newblock \bibinfo{title}{Domain-specific keyphrase extraction},
\newblock in: \bibinfo{editor}{T.~Dean} (Ed.), \bibinfo{booktitle}{Proceedings
  of the Sixteenth International Joint Conference on Artificial Intelligence,
  {IJCAI} 99, Stockholm, Sweden, July 31 - August 6, 1999. 2 Volumes, 1450
  pages}, \bibinfo{publisher}{Morgan Kaufmann}, \bibinfo{year}{1999}, pp.
  \bibinfo{pages}{668--673}.
\bibitem[{Gutwin et~al.(1999)Gutwin, Paynter, Witten, Nevill-Manning, and
  Frank}]{GUTWIN199981}
\bibinfo{author}{C.~Gutwin}, \bibinfo{author}{G.~Paynter},
  \bibinfo{author}{I.~Witten}, \bibinfo{author}{C.~Nevill-Manning},
  \bibinfo{author}{E.~Frank},
\newblock \bibinfo{title}{Improving browsing in digital libraries with
  keyphrase indexes},
\newblock \bibinfo{journal}{Decision Support Systems} \bibinfo{volume}{27}
  (\bibinfo{year}{1999}) \bibinfo{pages}{81--104}.
  \DOIprefix\doi{https://doi.org/10.1016/S0167-9236(99)00038-X}.
\bibitem[{Medelyan and Witten(2008)}]{10.5555/1364846.1364848}
\bibinfo{author}{O.~Medelyan}, \bibinfo{author}{I.~H. Witten},
\newblock \bibinfo{title}{Domain-independent automatic keyphrase indexing with
  small training sets},
\newblock \bibinfo{journal}{J. Am. Soc. Inf. Sci. Technol.}
  \bibinfo{volume}{59} (\bibinfo{year}{2008}) \bibinfo{pages}{1026–1040}.
\bibitem[{Borisov et~al.(2021)Borisov, Aliannejadi, and
  Crestani}]{DBLP:journals/corr/abs-2109-05979}
\bibinfo{author}{O.~Borisov}, \bibinfo{author}{M.~Aliannejadi},
  \bibinfo{author}{F.~Crestani},
\newblock \bibinfo{title}{Keyword extraction for improved document retrieval in
  conversational search},
\newblock \bibinfo{journal}{CoRR} \bibinfo{volume}{abs/2109.05979}
  (\bibinfo{year}{2021}).
\bibitem[{Han et~al.(2007)Han, Kim, and Choi}]{4427539}
\bibinfo{author}{J.~Han}, \bibinfo{author}{T.~Kim}, \bibinfo{author}{J.~Choi},
\newblock \bibinfo{title}{Web document clustering by using automatic keyphrase
  extraction},
\newblock in: \bibinfo{booktitle}{2007 IEEE/WIC/ACM International Conferences
  on Web Intelligence and Intelligent Agent Technology - Workshops},
  \bibinfo{year}{2007}, pp. \bibinfo{pages}{56--59}.
  \DOIprefix\doi{10.1109/WI-IATW.2007.46}.
\bibitem[{Hammouda et~al.(2005)Hammouda, Matute, and
  Kamel}]{10.1007/11510888_26}
\bibinfo{author}{K.~M. Hammouda}, \bibinfo{author}{D.~N. Matute},
  \bibinfo{author}{M.~S. Kamel},
\newblock \bibinfo{title}{Corephrase: Keyphrase extraction for document
  clustering},
\newblock in: \bibinfo{editor}{P.~Perner}, \bibinfo{editor}{A.~Imiya} (Eds.),
  \bibinfo{booktitle}{Machine Learning and Data Mining in Pattern Recognition},
  \bibinfo{publisher}{Springer Berlin Heidelberg}, \bibinfo{address}{Berlin,
  Heidelberg}, \bibinfo{year}{2005}, pp. \bibinfo{pages}{265--274}.
\bibitem[{Litvak and Last(2008)}]{10.5555/1613172.1613178}
\bibinfo{author}{M.~Litvak}, \bibinfo{author}{M.~Last},
\newblock \bibinfo{title}{Graph-based keyword extraction for single-document
  summarization},
\newblock in: \bibinfo{booktitle}{Proceedings of the Workshop on Multi-Source
  Multilingual Information Extraction and Summarization}, MMIES '08,
  \bibinfo{publisher}{Association for Computational Linguistics},
  \bibinfo{address}{USA}, \bibinfo{year}{2008}, p. \bibinfo{pages}{17–24}.
\bibitem[{Sarkar(2014)}]{10.4018/ijtd.2014040103}
\bibinfo{author}{K.~Sarkar},
\newblock \bibinfo{title}{A keyphrase-based approach to text summarization for
  english and bengali documents},
\newblock \bibinfo{journal}{Int. J. Technol. Diffus.} \bibinfo{volume}{5}
  (\bibinfo{year}{2014}) \bibinfo{pages}{28–38}.
  \DOIprefix\doi{10.4018/ijtd.2014040103}.
\bibitem[{Thomas et~al.(2016)Thomas, Bharti, and
  Babu}]{10.1145/2980258.2980442}
\bibinfo{author}{J.~R. Thomas}, \bibinfo{author}{S.~K. Bharti},
  \bibinfo{author}{K.~S. Babu},
\newblock \bibinfo{title}{Automatic keyword extraction for text summarization
  in e-newspapers},
\newblock in: \bibinfo{booktitle}{Proceedings of the International Conference
  on Informatics and Analytics}, ICIA-16, \bibinfo{publisher}{Association for
  Computing Machinery}, \bibinfo{address}{New York, NY, USA},
  \bibinfo{year}{2016}. \DOIprefix\doi{10.1145/2980258.2980442}.
\bibitem[{Kim et~al.(2010)Kim, Medelyan, Kan, and
  Baldwin}]{kim-etal-2010-semeval}
\bibinfo{author}{S.~N. Kim}, \bibinfo{author}{O.~Medelyan},
  \bibinfo{author}{M.-Y. Kan}, \bibinfo{author}{T.~Baldwin},
\newblock \bibinfo{title}{{S}em{E}val-2010 task 5 : Automatic keyphrase
  extraction from scientific articles},
\newblock in: \bibinfo{booktitle}{Proceedings of the 5th International Workshop
  on Semantic Evaluation}, \bibinfo{publisher}{Association for Computational
  Linguistics}, \bibinfo{address}{Uppsala, Sweden}, \bibinfo{year}{2010}, pp.
  \bibinfo{pages}{21--26}.
\bibitem[{Li et~al.(2021)Li, Li, and Xue}]{10.1007/978-981-15-8458-9_12}
\bibinfo{author}{J.~Li}, \bibinfo{author}{Y.~Li}, \bibinfo{author}{Z.~Xue},
\newblock \bibinfo{title}{Keywords extraction algorithm of financial review
  based on dirichlet multinomial model},
\newblock in: \bibinfo{editor}{Y.~Jia}, \bibinfo{editor}{W.~Zhang},
  \bibinfo{editor}{Y.~Fu} (Eds.), \bibinfo{booktitle}{Proceedings of 2020
  Chinese Intelligent Systems Conference}, \bibinfo{publisher}{Springer
  Singapore}, \bibinfo{address}{Singapore}, \bibinfo{year}{2021}, pp.
  \bibinfo{pages}{107--116}.
\bibitem[{Peji{\'{c}}~Bach et~al.(2019)Peji{\'{c}}~Bach, Krsti{\'{c}}, Seljan,
  and Turulja}]{su11051277}
\bibinfo{author}{M.~Peji{\'{c}}~Bach}, \bibinfo{author}{{\u{Z}}.~Krsti{\'{c}}},
  \bibinfo{author}{S.~Seljan}, \bibinfo{author}{L.~Turulja},
\newblock \bibinfo{title}{Text mining for big data analysis in financial
  sector: A literature review},
\newblock \bibinfo{journal}{Sustainability} \bibinfo{volume}{11}
  (\bibinfo{year}{2019}). \DOIprefix\doi{10.3390/su11051277}.
\bibitem[{Jungiewicz and
  {\L}opuszy{\'{n}}ski(2014)}]{10.1007/978-3-319-10888-9_7}
\bibinfo{author}{M.~Jungiewicz}, \bibinfo{author}{M.~{\L}opuszy{\'{n}}ski},
\newblock \bibinfo{title}{Unsupervised keyword extraction from polish legal
  texts},
\newblock in: \bibinfo{editor}{A.~Przepi{\'o}rkowski},
  \bibinfo{editor}{M.~Ogrodniczuk} (Eds.), \bibinfo{booktitle}{Advances in
  Natural Language Processing}, \bibinfo{publisher}{Springer International
  Publishing}, \bibinfo{address}{Cham}, \bibinfo{year}{2014}, pp.
  \bibinfo{pages}{65--70}.
\bibitem[{{Wu} et~al.(2022){Wu}, {Uddin Ahmad}, {Dev}, and
  {Chang}}]{2022arXiv220308118W}
\bibinfo{author}{D.~{Wu}}, \bibinfo{author}{W.~{Uddin Ahmad}},
  \bibinfo{author}{S.~{Dev}}, \bibinfo{author}{K.-W. {Chang}},
\newblock \bibinfo{title}{{Representation Learning for Resource-Constrained
  Keyphrase Generation}},
\newblock \bibinfo{journal}{arXiv e-prints}  (\bibinfo{year}{2022}).
  \href{http://arxiv.org/abs/2203.08118}{{\tt arXiv:2203.08118}}.
\bibitem[{Piskorski et~al.(2021)Piskorski, Stefanovitch, Jacquet, and
  Podavini}]{piskorski-etal-2021-exploring}
\bibinfo{author}{J.~Piskorski}, \bibinfo{author}{N.~Stefanovitch},
  \bibinfo{author}{G.~Jacquet}, \bibinfo{author}{A.~Podavini},
\newblock \bibinfo{title}{Exploring linguistically-lightweight keyword
  extraction techniques for indexing news articles in a multilingual set-up},
\newblock in: \bibinfo{booktitle}{Proceedings of the EACL Hackashop on News
  Media Content Analysis and Automated Report Generation},
  \bibinfo{publisher}{Association for Computational Linguistics},
  \bibinfo{address}{Online}, \bibinfo{year}{2021}, pp. \bibinfo{pages}{35--44}.
\bibitem[{Suzuki and Takatsuka(2016)}]{suzuki-takatsuka-2016-extraction}
\bibinfo{author}{S.~Suzuki}, \bibinfo{author}{H.~Takatsuka},
\newblock \bibinfo{title}{Extraction of keywords of novelties from patent
  claims},
\newblock in: \bibinfo{booktitle}{Proceedings of {COLING} 2016, the 26th
  International Conference on Computational Linguistics: Technical Papers},
  \bibinfo{publisher}{The COLING 2016 Organizing Committee},
  \bibinfo{address}{Osaka, Japan}, \bibinfo{year}{2016}, pp.
  \bibinfo{pages}{1192--1200}.
\bibitem[{Hu et~al.(2018)Hu, Li, Yao, Yu, Yang, and Hu}]{e20020104}
\bibinfo{author}{J.~Hu}, \bibinfo{author}{S.~Li}, \bibinfo{author}{Y.~Yao},
  \bibinfo{author}{L.~Yu}, \bibinfo{author}{G.~Yang}, \bibinfo{author}{J.~Hu},
\newblock \bibinfo{title}{Patent keyword extraction algorithm based on
  distributed representation for patent classification},
\newblock \bibinfo{journal}{Entropy} \bibinfo{volume}{20}
  (\bibinfo{year}{2018}). \DOIprefix\doi{10.3390/e20020104}.
\bibitem[{Ding and Luo(2021)}]{10.1145/3473939}
\bibinfo{author}{H.~Ding}, \bibinfo{author}{X.~Luo},
\newblock \bibinfo{title}{Attention-based unsupervised keyphrase extraction and
  phrase graph for covid-19 medical literature retrieval},
\newblock \bibinfo{journal}{ACM Trans. Comput. Healthcare} \bibinfo{volume}{3}
  (\bibinfo{year}{2021}). \DOIprefix\doi{10.1145/3473939}.
\bibitem[{Komenda et~al.(2016)Komenda, Karolyi, Pokorná, Víta, and
  Kríž}]{7733254}
\bibinfo{author}{M.~Komenda}, \bibinfo{author}{M.~Karolyi},
  \bibinfo{author}{A.~Pokorná}, \bibinfo{author}{M.~Víta},
  \bibinfo{author}{V.~Kríž},
\newblock \bibinfo{title}{Automatic keyword extraction from medical and
  healthcare curriculum},
\newblock in: \bibinfo{booktitle}{2016 Federated Conference on Computer Science
  and Information Systems (FedCSIS)}, \bibinfo{year}{2016}, pp.
  \bibinfo{pages}{287--290}.
\bibitem[{Li and Wu(2006)}]{LI2006668}
\bibinfo{author}{Q.~Li}, \bibinfo{author}{Y.-F.~B. Wu},
\newblock \bibinfo{title}{Identifying important concepts from medical
  documents},
\newblock \bibinfo{journal}{Journal of Biomedical Informatics}
  \bibinfo{volume}{39} (\bibinfo{year}{2006}) \bibinfo{pages}{668--679}.
  \DOIprefix\doi{https://doi.org/10.1016/j.jbi.2006.02.001}.
\bibitem[{Zehtab-Salmasi et~al.(2021)Zehtab-Salmasi, Feizi-Derakhshi, and
  Balafar}]{zehtabsalmasi2021frake}
\bibinfo{author}{A.~Zehtab-Salmasi}, \bibinfo{author}{M.-R. Feizi-Derakhshi},
  \bibinfo{author}{M.-A. Balafar}, \bibinfo{title}{Frake: Fusional real-time
  automatic keyword extraction}, \bibinfo{year}{2021}.
  \href{http://arxiv.org/abs/2104.04830}{{\tt arXiv:2104.04830}}.
\bibitem[{Sun et~al.(2020)Sun, Hu, Li, Li, Li, and Chi}]{sym12111864}
\bibinfo{author}{C.~Sun}, \bibinfo{author}{L.~Hu}, \bibinfo{author}{S.~Li},
  \bibinfo{author}{T.~Li}, \bibinfo{author}{H.~Li}, \bibinfo{author}{L.~Chi},
\newblock \bibinfo{title}{A review of unsupervised keyphrase extraction methods
  using within-collection resources},
\newblock \bibinfo{journal}{Symmetry} \bibinfo{volume}{12}
  (\bibinfo{year}{2020}). \DOIprefix\doi{10.3390/sym12111864}.
\bibitem[{Mahata et~al.(2018)Mahata, Shah, Kuriakose, Zimmermann, and
  Talburt}]{8396997}
\bibinfo{author}{D.~Mahata}, \bibinfo{author}{R.~R. Shah},
  \bibinfo{author}{J.~Kuriakose}, \bibinfo{author}{R.~Zimmermann},
  \bibinfo{author}{J.~R. Talburt},
\newblock \bibinfo{title}{Theme-weighted ranking of keywords from text
  documents using phrase embeddings},
\newblock in: \bibinfo{booktitle}{2018 IEEE Conference on Multimedia
  Information Processing and Retrieval (MIPR)}, \bibinfo{year}{2018}, pp.
  \bibinfo{pages}{184--189}. \DOIprefix\doi{10.1109/MIPR.2018.00041}.
\bibitem[{Kuai et~al.(2021)Kuai, Liao, Chang, and Yu}]{9602870}
\bibinfo{author}{S.-C. Kuai}, \bibinfo{author}{W.-H. Liao},
  \bibinfo{author}{C.-Y. Chang}, \bibinfo{author}{G.-J. Yu},
\newblock \bibinfo{title}{Fb-kea: A feature-based keyword extraction algorithm
  for improving hit performance},
\newblock in: \bibinfo{booktitle}{2021 IEEE International Conference on
  Consumer Electronics-Taiwan (ICCE-TW)}, \bibinfo{year}{2021}, pp.
  \bibinfo{pages}{1--2}. \DOIprefix\doi{10.1109/ICCE-TW52618.2021.9602870}.
\bibitem[{Saga et~al.(2014)Saga, Kobayashi, Miyamoto, and
  Tsuji}]{10.1007/978-3-319-07857-1_40}
\bibinfo{author}{R.~Saga}, \bibinfo{author}{H.~Kobayashi},
  \bibinfo{author}{T.~Miyamoto}, \bibinfo{author}{H.~Tsuji},
\newblock \bibinfo{title}{Measurement evaluation of keyword extraction based on
  topic coverage},
\newblock in: \bibinfo{editor}{C.~Stephanidis} (Ed.), \bibinfo{booktitle}{HCI
  International 2014 - Posters' Extended Abstracts},
  \bibinfo{publisher}{Springer International Publishing},
  \bibinfo{address}{Cham}, \bibinfo{year}{2014}, pp. \bibinfo{pages}{224--227}.
\bibitem[{Liu et~al.(2020)Liu, Huang, Huang, and Duan}]{sym12111923}
\bibinfo{author}{F.~Liu}, \bibinfo{author}{X.~Huang},
  \bibinfo{author}{W.~Huang}, \bibinfo{author}{S.~X. Duan},
\newblock \bibinfo{title}{Performance evaluation of keyword extraction methods
  and visualization for student online comments},
\newblock \bibinfo{journal}{Symmetry} \bibinfo{volume}{12}
  (\bibinfo{year}{2020}). \DOIprefix\doi{10.3390/sym12111923}.
\bibitem[{Bornmann and Mutz(5 11)}]{20.500.11850/106263}
\bibinfo{author}{L.~Bornmann}, \bibinfo{author}{R.~Mutz},
\newblock \bibinfo{title}{Growth rates of modern science: A bibliometric
  analysis based on the number of publications and cited references},
\newblock \bibinfo{journal}{Journal of the Association for Information Science
  and Technology : JASIST} \bibinfo{volume}{66} (\bibinfo{year}{2015-11})
  \bibinfo{pages}{2215 -- 2222}. \DOIprefix\doi{10.1002/asi.23329},
  \bibinfo{note}{published online 29 April 2015.}
\bibitem[{Hua and Shin(2021)}]{DBLP:conf/iconference/HuaS21}
\bibinfo{author}{B.~Hua}, \bibinfo{author}{Y.~Shin},
\newblock \bibinfo{title}{Extraction of sentences describing originality from
  conclusion in academic papers},
\newblock in: \bibinfo{editor}{Y.~Zhang}, \bibinfo{editor}{C.~Zhang},
  \bibinfo{editor}{P.~Mayr}, \bibinfo{editor}{A.~Suominen} (Eds.),
  \bibinfo{booktitle}{Proceedings of the 1st Workshop on {AI} + Informetrics
  {(AII2021)} co-located with the iConference 2021, Virtual Event, March 17th,
  2021}, volume \bibinfo{volume}{2871} of \textit{\bibinfo{series}{{CEUR}
  Workshop Proceedings}}, \bibinfo{publisher}{CEUR-WS.org},
  \bibinfo{year}{2021}, pp. \bibinfo{pages}{58--70}.
\bibitem[{Krallinger et~al.(2015)Krallinger, Leitner, Rabal, Vazquez,
  Oyarz{\'a}bal, and Valencia}]{Krallinger2015CHEMDNERTD}
\bibinfo{author}{M.~Krallinger}, \bibinfo{author}{F.~Leitner},
  \bibinfo{author}{O.~Rabal}, \bibinfo{author}{M.~Vazquez},
  \bibinfo{author}{J.~Oyarz{\'a}bal}, \bibinfo{author}{A.~Valencia},
\newblock \bibinfo{title}{Chemdner: The drugs and chemical names extraction
  challenge},
\newblock \bibinfo{journal}{Journal of Cheminformatics} \bibinfo{volume}{7}
  (\bibinfo{year}{2015}) \bibinfo{pages}{S1 -- S1}.
\bibitem[{Porter(1997)}]{10.5555/275537.275705}
\bibinfo{author}{M.~F. Porter}, \bibinfo{title}{An Algorithm for Suffix
  Stripping}, \bibinfo{publisher}{Morgan Kaufmann Publishers Inc.},
  \bibinfo{address}{San Francisco, CA, USA}, \bibinfo{year}{1997}, p.
  \bibinfo{pages}{313–316}.
\bibitem[{Grootendorst(2020)}]{grootendorst2020keybert}
\bibinfo{author}{M.~Grootendorst}, \bibinfo{title}{Keybert: Minimal keyword
  extraction with bert.}, \bibinfo{year}{2020}.
  \DOIprefix\doi{10.5281/zenodo.4461265}.
\bibitem[{Kowsari et~al.(2017)Kowsari, Brown, Heidarysafa, Jafari~Meimandi, ,
  Gerber, and Barnes}]{kowsari2017HDLTex}
\bibinfo{author}{K.~Kowsari}, \bibinfo{author}{D.~E. Brown},
  \bibinfo{author}{M.~Heidarysafa}, \bibinfo{author}{K.~Jafari~Meimandi}, ,
  \bibinfo{author}{M.~S. Gerber}, \bibinfo{author}{L.~E. Barnes},
\newblock \bibinfo{title}{Hdltex: Hierarchical deep learning for text
  classification},
\newblock in: \bibinfo{booktitle}{Machine Learning and Applications (ICMLA),
  2017 16th IEEE International Conference on}, \bibinfo{organization}{IEEE},
  \bibinfo{year}{2017}.
\bibitem[{Mihalcea and Tarau(2004)}]{mihalcea-tarau-2004-textrank}
\bibinfo{author}{R.~Mihalcea}, \bibinfo{author}{P.~Tarau},
\newblock \bibinfo{title}{{T}ext{R}ank: Bringing order into text},
\newblock in: \bibinfo{booktitle}{Proceedings of the 2004 Conference on
  Empirical Methods in Natural Language Processing},
  \bibinfo{publisher}{Association for Computational Linguistics},
  \bibinfo{address}{Barcelona, Spain}, \bibinfo{year}{2004}, pp.
  \bibinfo{pages}{404--411}.
\bibitem[{Lloyd(1982)}]{1056489}
\bibinfo{author}{S.~Lloyd},
\newblock \bibinfo{title}{Least squares quantization in pcm},
\newblock \bibinfo{journal}{IEEE Transactions on Information Theory}
  \bibinfo{volume}{28} (\bibinfo{year}{1982}) \bibinfo{pages}{129--137}.
  \DOIprefix\doi{10.1109/TIT.1982.1056489}.
\bibitem[{MacQueen(1967)}]{macqueen1967classification}
\bibinfo{author}{J.~MacQueen},
\newblock \bibinfo{title}{Classification and analysis of multivariate
  observations},
\newblock in: \bibinfo{booktitle}{5th Berkeley Symp. Math. Statist.
  Probability}, \bibinfo{year}{1967}, pp. \bibinfo{pages}{281--297}.
\bibitem[{Weber et~al.(2021)Weber, Sänger, Münchmeyer, Habibi, Leser, and
  Akbik}]{10.1093/bioinformatics/btab042}
\bibinfo{author}{L.~Weber}, \bibinfo{author}{M.~Sänger},
  \bibinfo{author}{J.~Münchmeyer}, \bibinfo{author}{M.~Habibi},
  \bibinfo{author}{U.~Leser}, \bibinfo{author}{A.~Akbik},
\newblock \bibinfo{title}{{HunFlair: an easy-to-use tool for state-of-the-art
  biomedical named entity recognition}},
\newblock \bibinfo{journal}{Bioinformatics} \bibinfo{volume}{37}
  (\bibinfo{year}{2021}) \bibinfo{pages}{2792--2794}.
  \DOIprefix\doi{10.1093/bioinformatics/btab042}.
\bibitem[{Son and Shin(2018)}]{ART002312659}
\bibinfo{author}{J.~Son}, \bibinfo{author}{Y.~Shin},
\newblock \bibinfo{title}{Music lyrics summarization method using textrank
  algorithm},
\newblock \bibinfo{journal}{Journal of Korea Multimedia Society}
  \bibinfo{volume}{21} (\bibinfo{year}{2018}) \bibinfo{pages}{45--50}.
  \DOIprefix\doi{https://doi.org/10.9717/kmms.2018.21.1.045}.
\bibitem[{Wu et~al.(2022)Wu, Liao, Afedzie~Kwofie, Zou, Wang, and
  Zhang}]{10.1007/978-981-16-2377-6_46}
\bibinfo{author}{C.~Wu}, \bibinfo{author}{L.~Liao},
  \bibinfo{author}{F.~Afedzie~Kwofie}, \bibinfo{author}{F.~Zou},
  \bibinfo{author}{Y.~Wang}, \bibinfo{author}{M.~Zhang},
\newblock \bibinfo{title}{Textrank keyword extraction method based on
  multi-feature fusion},
\newblock in: \bibinfo{editor}{X.-S. Yang}, \bibinfo{editor}{S.~Sherratt},
  \bibinfo{editor}{N.~Dey}, \bibinfo{editor}{A.~Joshi} (Eds.),
  \bibinfo{booktitle}{Proceedings of Sixth International Congress on
  Information and Communication Technology}, \bibinfo{publisher}{Springer
  Singapore}, \bibinfo{address}{Singapore}, \bibinfo{year}{2022}, pp.
  \bibinfo{pages}{493--501}.
\bibitem[{Pan et~al.(2019)Pan, Li, and Dai}]{10.1145/3321408.3326659}
\bibinfo{author}{S.~Pan}, \bibinfo{author}{Z.~Li}, \bibinfo{author}{J.~Dai},
\newblock \bibinfo{title}{An improved textrank keywords extraction algorithm},
\newblock in: \bibinfo{booktitle}{Proceedings of the ACM Turing Celebration
  Conference - China}, ACM TURC '19, \bibinfo{publisher}{Association for
  Computing Machinery}, \bibinfo{address}{New York, NY, USA},
  \bibinfo{year}{2019}. \DOIprefix\doi{10.1145/3321408.3326659}.
\bibitem[{Montani et~al.(2022)Montani, Honnibal, Honnibal, Landeghem, Boyd,
  Peters, McCann, Samsonov, Geovedi, O'Regan, Altinok, Orosz, Kristiansen,
  de~Kok, Miranda, Roman, Bot, Fiedler, Howard, Edward, Phatthiyaphaibun,
  Hudson, Tamura, Bozek, murat, Daniels, Baumgartner, Amery, and
  Böing}]{ines_montani_2022_6621076}
\bibinfo{author}{I.~Montani}, \bibinfo{author}{M.~Honnibal},
  \bibinfo{author}{M.~Honnibal}, \bibinfo{author}{S.~V. Landeghem},
  \bibinfo{author}{A.~Boyd}, \bibinfo{author}{H.~Peters},
  \bibinfo{author}{P.~O. McCann}, \bibinfo{author}{M.~Samsonov},
  \bibinfo{author}{J.~Geovedi}, \bibinfo{author}{J.~O'Regan},
  \bibinfo{author}{D.~Altinok}, \bibinfo{author}{G.~Orosz},
  \bibinfo{author}{S.~L. Kristiansen}, \bibinfo{author}{D.~de~Kok},
  \bibinfo{author}{L.~Miranda}, \bibinfo{author}{Roman},
  \bibinfo{author}{E.~Bot}, \bibinfo{author}{L.~Fiedler},
  \bibinfo{author}{G.~Howard}, \bibinfo{author}{Edward},
  \bibinfo{author}{W.~Phatthiyaphaibun}, \bibinfo{author}{R.~Hudson},
  \bibinfo{author}{Y.~Tamura}, \bibinfo{author}{S.~Bozek},
  \bibinfo{author}{murat}, \bibinfo{author}{R.~Daniels},
  \bibinfo{author}{P.~Baumgartner}, \bibinfo{author}{M.~Amery},
  \bibinfo{author}{B.~Böing}, \bibinfo{title}{{explosion/spaCy: New Span Ruler
  component, JSON (de)serialization of Doc, span analyzer and more}},
  \bibinfo{year}{2022}. \DOIprefix\doi{10.5281/zenodo.6621076}.
\bibitem[{Barrios et~al.(2016)Barrios, L{\'{o}}pez, Argerich, and
  Wachenchauzer}]{DBLP:journals/corr/BarriosLAW16}
\bibinfo{author}{F.~Barrios}, \bibinfo{author}{F.~L{\'{o}}pez},
  \bibinfo{author}{L.~Argerich}, \bibinfo{author}{R.~Wachenchauzer},
\newblock \bibinfo{title}{Variations of the similarity function of textrank for
  automated summarization},
\newblock \bibinfo{journal}{CoRR} \bibinfo{volume}{abs/1602.03606}
  (\bibinfo{year}{2016}). \href{http://arxiv.org/abs/1602.03606}{{\tt
  arXiv:1602.03606}}.
\bibitem[{Boudin(2016)}]{boudin:2016:COLINGDEMO}
\bibinfo{author}{F.~Boudin},
\newblock \bibinfo{title}{pke: an open source python-based keyphrase extraction
  toolkit},
\newblock in: \bibinfo{booktitle}{Proceedings of COLING 2016, the 26th
  International Conference on Computational Linguistics: System
  Demonstrations}, \bibinfo{address}{Osaka, Japan}, \bibinfo{year}{2016}, pp.
  \bibinfo{pages}{69--73}.
\bibitem[{Neumann et~al.(2019)Neumann, King, Beltagy, and
  Ammar}]{neumann-etal-2019-scispacy}
\bibinfo{author}{M.~Neumann}, \bibinfo{author}{D.~King},
  \bibinfo{author}{I.~Beltagy}, \bibinfo{author}{W.~Ammar},
\newblock \bibinfo{title}{{S}cispa{C}y: Fast and robust models for biomedical
  natural language processing},
\newblock in: \bibinfo{booktitle}{Proceedings of the 18th BioNLP Workshop and
  Shared Task}, \bibinfo{publisher}{Association for Computational Linguistics},
  \bibinfo{address}{Florence, Italy}, \bibinfo{year}{2019}, pp.
  \bibinfo{pages}{319--327}. \DOIprefix\doi{10.18653/v1/W19-5034}.
\bibitem[{Hagberg et~al.(2008)Hagberg, Swart, and S~Chult}]{osti_960616}
\bibinfo{author}{A.~Hagberg}, \bibinfo{author}{P.~Swart},
  \bibinfo{author}{D.~S~Chult},
\newblock \bibinfo{title}{Exploring network structure, dynamics, and function
  using networkx}  (\bibinfo{year}{2008}).
\bibitem[{Xu and Tian(2015)}]{Xu2015}
\bibinfo{author}{D.~Xu}, \bibinfo{author}{Y.~Tian},
\newblock \bibinfo{title}{A comprehensive survey of clustering algorithms},
\newblock \bibinfo{journal}{Annals of Data Science} \bibinfo{volume}{2}
  (\bibinfo{year}{2015}) \bibinfo{pages}{165--193}.
  \DOIprefix\doi{10.1007/s40745-015-0040-1}.
\bibitem[{Ezugwu et~al.(2022)Ezugwu, Ikotun, Oyelade, Abualigah, Agushaka, Eke,
  and Akinyelu}]{EZUGWU2022104743}
\bibinfo{author}{A.~E. Ezugwu}, \bibinfo{author}{A.~M. Ikotun},
  \bibinfo{author}{O.~O. Oyelade}, \bibinfo{author}{L.~Abualigah},
  \bibinfo{author}{J.~O. Agushaka}, \bibinfo{author}{C.~I. Eke},
  \bibinfo{author}{A.~A. Akinyelu},
\newblock \bibinfo{title}{A comprehensive survey of clustering algorithms:
  State-of-the-art machine learning applications, taxonomy, challenges, and
  future research prospects},
\newblock \bibinfo{journal}{Engineering Applications of Artificial
  Intelligence} \bibinfo{volume}{110} (\bibinfo{year}{2022})
  \bibinfo{pages}{104743}.
  \DOIprefix\doi{https://doi.org/10.1016/j.engappai.2022.104743}.
\bibitem[{Pennington et~al.(2014)Pennington, Socher, and
  Manning}]{pennington-etal-2014-glove}
\bibinfo{author}{J.~Pennington}, \bibinfo{author}{R.~Socher},
  \bibinfo{author}{C.~Manning},
\newblock \bibinfo{title}{{G}lo{V}e: Global vectors for word representation},
\newblock in: \bibinfo{booktitle}{Proceedings of the 2014 Conference on
  Empirical Methods in Natural Language Processing ({EMNLP})},
  \bibinfo{publisher}{Association for Computational Linguistics},
  \bibinfo{address}{Doha, Qatar}, \bibinfo{year}{2014}, pp.
  \bibinfo{pages}{1532--1543}. \DOIprefix\doi{10.3115/v1/D14-1162}.
\bibitem[{Devlin et~al.(2019)Devlin, Chang, Lee, and
  Toutanova}]{devlin-etal-2019-bert}
\bibinfo{author}{J.~Devlin}, \bibinfo{author}{M.-W. Chang},
  \bibinfo{author}{K.~Lee}, \bibinfo{author}{K.~Toutanova},
\newblock \bibinfo{title}{{BERT}: Pre-training of deep bidirectional
  transformers for language understanding},
\newblock in: \bibinfo{booktitle}{Proceedings of the 2019 Conference of the
  North {A}merican Chapter of the Association for Computational Linguistics:
  Human Language Technologies, Volume 1 (Long and Short Papers)},
  \bibinfo{publisher}{Association for Computational Linguistics},
  \bibinfo{address}{Minneapolis, Minnesota}, \bibinfo{year}{2019}, pp.
  \bibinfo{pages}{4171--4186}. \DOIprefix\doi{10.18653/v1/N19-1423}.
\bibitem[{Reimers and Gurevych(2019)}]{reimers-gurevych-2019-sentence}
\bibinfo{author}{N.~Reimers}, \bibinfo{author}{I.~Gurevych},
\newblock \bibinfo{title}{Sentence-{BERT}: Sentence embeddings using {S}iamese
  {BERT}-networks},
\newblock in: \bibinfo{booktitle}{Proceedings of the 2019 Conference on
  Empirical Methods in Natural Language Processing and the 9th International
  Joint Conference on Natural Language Processing (EMNLP-IJCNLP)},
  \bibinfo{publisher}{Association for Computational Linguistics},
  \bibinfo{address}{Hong Kong, China}, \bibinfo{year}{2019}, pp.
  \bibinfo{pages}{3982--3992}. \DOIprefix\doi{10.18653/v1/D19-1410}.
\bibitem[{Pedregosa et~al.(2011)Pedregosa, Varoquaux, Gramfort, Michel,
  Thirion, Grisel, Blondel, Prettenhofer, Weiss, Dubourg, Vanderplas, Passos,
  Cournapeau, Brucher, Perrot, and Duchesnay}]{scikit-learn}
\bibinfo{author}{F.~Pedregosa}, \bibinfo{author}{G.~Varoquaux},
  \bibinfo{author}{A.~Gramfort}, \bibinfo{author}{V.~Michel},
  \bibinfo{author}{B.~Thirion}, \bibinfo{author}{O.~Grisel},
  \bibinfo{author}{M.~Blondel}, \bibinfo{author}{P.~Prettenhofer},
  \bibinfo{author}{R.~Weiss}, \bibinfo{author}{V.~Dubourg},
  \bibinfo{author}{J.~Vanderplas}, \bibinfo{author}{A.~Passos},
  \bibinfo{author}{D.~Cournapeau}, \bibinfo{author}{M.~Brucher},
  \bibinfo{author}{M.~Perrot}, \bibinfo{author}{E.~Duchesnay},
\newblock \bibinfo{title}{Scikit-learn: Machine learning in {P}ython},
\newblock \bibinfo{journal}{Journal of Machine Learning Research}
  \bibinfo{volume}{12} (\bibinfo{year}{2011}) \bibinfo{pages}{2825--2830}.
\bibitem[{Hunter(2007)}]{Hunter:2007}
\bibinfo{author}{J.~D. Hunter},
\newblock \bibinfo{title}{Matplotlib: A 2d graphics environment},
\newblock \bibinfo{journal}{Computing in Science \& Engineering}
  \bibinfo{volume}{9} (\bibinfo{year}{2007}) \bibinfo{pages}{90--95}.
  \DOIprefix\doi{10.1109/MCSE.2007.55}.
\bibitem[{Bougouin et~al.(2013)Bougouin, Boudin, and
  Daille}]{bougouin:hal-00917969}
\bibinfo{author}{A.~Bougouin}, \bibinfo{author}{F.~Boudin},
  \bibinfo{author}{B.~Daille},
\newblock \bibinfo{title}{{TopicRank: Graph-Based Topic Ranking for Keyphrase
  Extraction}},
\newblock in: \bibinfo{booktitle}{{International Joint Conference on Natural
  Language Processing (IJCNLP)}}, \bibinfo{address}{Nagoya, Japan},
  \bibinfo{year}{2013}, pp. \bibinfo{pages}{543--551}. \URLprefix
  \url{https://hal.archives-ouvertes.fr/hal-00917969}.
\bibitem[{Boudin(2018)}]{boudin-2018-unsupervised}
\bibinfo{author}{F.~Boudin},
\newblock \bibinfo{title}{Unsupervised keyphrase extraction with multipartite
  graphs},
\newblock in: \bibinfo{booktitle}{Proceedings of the 2018 Conference of the
  North {A}merican Chapter of the Association for Computational Linguistics:
  Human Language Technologies, Volume 2 (Short Papers)},
  \bibinfo{publisher}{Association for Computational Linguistics},
  \bibinfo{address}{New Orleans, Louisiana}, \bibinfo{year}{2018}, pp.
  \bibinfo{pages}{667--672}. \DOIprefix\doi{10.18653/v1/N18-2105}.
\bibitem[{Grootendorst(2022)}]{grootendorst2022bertopic}
\bibinfo{author}{M.~Grootendorst},
\newblock \bibinfo{title}{Bertopic: Neural topic modeling with a class-based
  tf-idf procedure},
\newblock \bibinfo{journal}{arXiv preprint arXiv:2203.05794}
  (\bibinfo{year}{2022}).
\bibitem[{Jurafsky and Martin(2018)}]{jurafsky2018speech}
\bibinfo{author}{D.~Jurafsky}, \bibinfo{author}{J.~H. Martin},
\newblock \bibinfo{title}{Speech and language processing (draft)},
\newblock \bibinfo{journal}{preparation [cited 2020 June 1] Available from:
  https://web. stanford. edu/\~{} jurafsky/slp3}  (\bibinfo{year}{2018}).
\bibitem[{Berry and Kogan(2010)}]{berry2010text}
\bibinfo{author}{M.~W. Berry}, \bibinfo{author}{J.~Kogan}, \bibinfo{title}{Text
  mining: applications and theory}, \bibinfo{publisher}{John Wiley \& Sons},
  \bibinfo{year}{2010}.
\bibitem[{Mansouri et~al.(2008)Mansouri, Affendey, and
  Mamat}]{mansouri2008named}
\bibinfo{author}{A.~Mansouri}, \bibinfo{author}{L.~S. Affendey},
  \bibinfo{author}{A.~Mamat},
\newblock \bibinfo{title}{Named entity recognition approaches},
\newblock \bibinfo{journal}{International Journal of Computer Science and
  Network Security} \bibinfo{volume}{8} (\bibinfo{year}{2008})
  \bibinfo{pages}{339--344}.
\bibitem[{Akbik et~al.(2019)Akbik, Bergmann, Blythe, Rasul, Schweter, and
  Vollgraf}]{akbik2019flair}
\bibinfo{author}{A.~Akbik}, \bibinfo{author}{T.~Bergmann},
  \bibinfo{author}{D.~Blythe}, \bibinfo{author}{K.~Rasul},
  \bibinfo{author}{S.~Schweter}, \bibinfo{author}{R.~Vollgraf},
\newblock \bibinfo{title}{{FLAIR}: An easy-to-use framework for
  state-of-the-art {NLP}},
\newblock in: \bibinfo{booktitle}{{NAACL} 2019, 2019 Annual Conference of the
  North American Chapter of the Association for Computational Linguistics
  (Demonstrations)}, \bibinfo{year}{2019}, pp. \bibinfo{pages}{54--59}.
\bibitem[{Neumann et~al.(2019)Neumann, King, Beltagy, and
  Ammar}]{DBLP:journals/corr/abs-1902-07669}
\bibinfo{author}{M.~Neumann}, \bibinfo{author}{D.~King},
  \bibinfo{author}{I.~Beltagy}, \bibinfo{author}{W.~Ammar},
\newblock \bibinfo{title}{Scispacy: Fast and robust models for biomedical
  natural language processing},
\newblock \bibinfo{journal}{CoRR} \bibinfo{volume}{abs/1902.07669}
  (\bibinfo{year}{2019}).
\bibitem[{Akbik et~al.(2018)Akbik, Blythe, and Vollgraf}]{akbik2018coling}
\bibinfo{author}{A.~Akbik}, \bibinfo{author}{D.~Blythe},
  \bibinfo{author}{R.~Vollgraf},
\newblock \bibinfo{title}{Contextual string embeddings for sequence labeling},
\newblock in: \bibinfo{booktitle}{{COLING} 2018, 27th International Conference
  on Computational Linguistics}, \bibinfo{year}{2018}, pp.
  \bibinfo{pages}{1638--1649}.
\bibitem[{Tjong Kim~Sang and
  De~Meulder(2003)}]{tjong-kim-sang-de-meulder-2003-introduction}
\bibinfo{author}{E.~F. Tjong Kim~Sang}, \bibinfo{author}{F.~De~Meulder},
\newblock \bibinfo{title}{Introduction to the {C}o{NLL}-2003 shared task:
  Language-independent named entity recognition},
\newblock in: \bibinfo{booktitle}{Proceedings of the Seventh Conference on
  Natural Language Learning at {HLT}-{NAACL} 2003}, \bibinfo{year}{2003}, pp.
  \bibinfo{pages}{142--147}.
\bibitem[{Weber et~al.(2019)Weber, Münchmeyer, Rocktäschel, Habibi, and
  Leser}]{10.1093/bioinformatics/btz528}
\bibinfo{author}{L.~Weber}, \bibinfo{author}{J.~Münchmeyer},
  \bibinfo{author}{T.~Rocktäschel}, \bibinfo{author}{M.~Habibi},
  \bibinfo{author}{U.~Leser},
\newblock \bibinfo{title}{{HUNER: improving biomedical NER with pretraining}},
\newblock \bibinfo{journal}{Bioinformatics} \bibinfo{volume}{36}
  (\bibinfo{year}{2019}) \bibinfo{pages}{295--302}.
  \DOIprefix\doi{10.1093/bioinformatics/btz528}.
\bibitem[{Marrero et~al.(2013)Marrero, Urbano, Sánchez-Cuadrado, Morato, and
  Gómez-Berbís}]{MARRERO2013482}
\bibinfo{author}{M.~Marrero}, \bibinfo{author}{J.~Urbano},
  \bibinfo{author}{S.~Sánchez-Cuadrado}, \bibinfo{author}{J.~Morato},
  \bibinfo{author}{J.~M. Gómez-Berbís},
\newblock \bibinfo{title}{Named entity recognition: Fallacies, challenges and
  opportunities},
\newblock \bibinfo{journal}{Computer Standards \& Interfaces}
  \bibinfo{volume}{35} (\bibinfo{year}{2013}) \bibinfo{pages}{482--489}.
  \DOIprefix\doi{https://doi.org/10.1016/j.csi.2012.09.004}.

\end{thebibliography}

\pagebreak

\appendix
\section{List of participants to the workshop}
We thank our workshop participants for valuable feedback, contributions, and suggestions. \\ \\
\small
\indent \textbf{Susie Xi Rao}, ETH Zurich (Organizer) \\
\indent \textbf{Piriyakorn Piriyatamwong}, ETH Zurich (Organizer) \\
\indent \textbf{Parijat Ghoshal}, NZZ AG (Organizer) \\
\indent Vanya Brucker, Wyona AG \\
\indent Andrea Bussolan, SUPSI \\
\indent Mercedes Garc\'{i}a Mart\'{i}nez, Pangeanic \\
\indent Sandra Mitrovi\'{c}, IDSIA USI-SUPSI \\
\indent Sara Nasirian, SUPSI \\
\indent Emmanuel de Salis, HE-Arc \\
\indent Natasa Sarafijanovic-Djukic, FFHS \\
\indent Dietrich Trautmann, Thomson Reuters \\
\indent Michael Wechner, Wyona AG \\
\indent Peter Egger, ETH Zurich (Principal Investigator) \\
\indent Ce Zhang, ETH Zurich (Principal Investigator) \\

\end{document}